%% file: neurips_2024.tex
\documentclass{article}

\usepackage[dvipsnames]{xcolor}

\usepackage[final, nonatbib]{neurips_2024}

\usepackage[utf8]{inputenc} 
\usepackage[T1]{fontenc}    
\usepackage{url}            
\usepackage{booktabs}       
\usepackage{amsfonts}       
\usepackage{nicefrac}       
\usepackage{microtype}      
\usepackage{xcolor}         
\usepackage[utf8]{inputenc} 
\usepackage[T1]{fontenc}    
\usepackage{url}            
\usepackage{booktabs}       
\usepackage{amsfonts}       
\usepackage{nicefrac}       
\usepackage{microtype}      
\usepackage{tabularx}
\usepackage{enumitem}
\usepackage[misc]{ifsym}
\usepackage{pdfpages}

\usepackage{color, colortbl}

\usepackage{algorithm}
\usepackage{listings}

\lstdefinelanguage{CUDA}{
  keywords={__global__, __device__, __host__, threadIdx, blockIdx, blockDim, __shared__},
  keywordstyle=\ttfamily\color{green!70!black},
  morekeywords={int, float, double, char, void, template, typename},
  identifierstyle=\ttfamily\color{blue!70!black},
  sensitive=true,
  comment=[l]{//},
  morecomment=[s]{/*}{*/},
  stringstyle=\ttfamily\color{red!70!black},
  showstringspaces=false,
  basicstyle=\footnotesize\ttfamily,
  backgroundcolor=\color{white},
  frame=none,
  framerule=0.5pt,
  rulecolor=\color{gray!70}
}

\usepackage{caption}
\usepackage{float}
\definecolor{DGray}{gray}{0.45}
\definecolor{Gray}{gray}{0.9}
\definecolor{codeblue}{rgb}{0.25,0.5,0.5}
\definecolor{codekw}{rgb}{0.85, 0.18, 0.50}
\definecolor{codekwb}{rgb}{0.0, 0.0, 1.00}
\floatstyle{plain}
\newfloat{Code}{htbp}{Code}
\restylefloat*{Code}
\floatname{Code}{Code}
\usepackage{amsmath}
\usepackage[pagebackref=true,breaklinks=true,colorlinks,bookmarks=false]{hyperref}
\usepackage[capitalize]{cleveref}
\crefname{section}{Sec.}{Secs.}
\Crefname{section}{Section}{Sections}
\Crefname{table}{Table}{Tables}
\crefname{table}{Tab.}{Tabs.}

\usepackage{times}
\usepackage{epsfig}
\usepackage{graphicx}
\usepackage{amssymb}

\usepackage{multirow}
\usepackage{booktabs}
\usepackage{float}
\usepackage{makecell}  
\usepackage{diagbox} 
\usepackage{multirow}
\usepackage{subfig}
\usepackage{bbding}
\usepackage{adjustbox}
\usepackage{pifont}

\newlength\savewidth\newcommand\shline{\noalign{\global\savewidth\arrayrulewidth
  \global\arrayrulewidth 1pt}\hline\noalign{\global\arrayrulewidth\savewidth}}

\newcolumntype{x}[1]{>{\centering\arraybackslash}p{#1pt}}
\newcolumntype{y}[1]{>{\raggedright\arraybackslash}p{#1pt}}
\newcolumntype{z}[1]{>{\raggedleft\arraybackslash}p{#1pt}}
\renewcommand{\paragraph}[1]{\vspace{1.25mm}\noindent\textbf{#1}}
\definecolor{deemph}{gray}{0.6}

\newcommand{\eg}{\textit{e.g.}}

\title{FlowDCN: Exploring DCN-like Architectures for Fast Image Generation with Arbitrary Resolution}

\author{
Shuai Wang \\
Nanjing University \\
\And
Zexian Li \\
Alibaba Group \\
\And
Tianhui Song \\
Nanjing University \\
\And 
Xubin Li \\
Alibaba Group \\
\And
Tiezheng Ge \\
Alibaba Group \\
\And 
Bo Zheng \\
Alibaba Group \\
\And 
Limin Wang \Letter \\ 
Nanjing University, Shanghai AI Lab \\
}

\begin{document}

\renewcommand\twocolumn[1][]{#1}%
\maketitle
\vspace{-8mm}
\begin{center}
    \centering
    \captionsetup{type=figure}
    \includegraphics[width=0.90\linewidth]{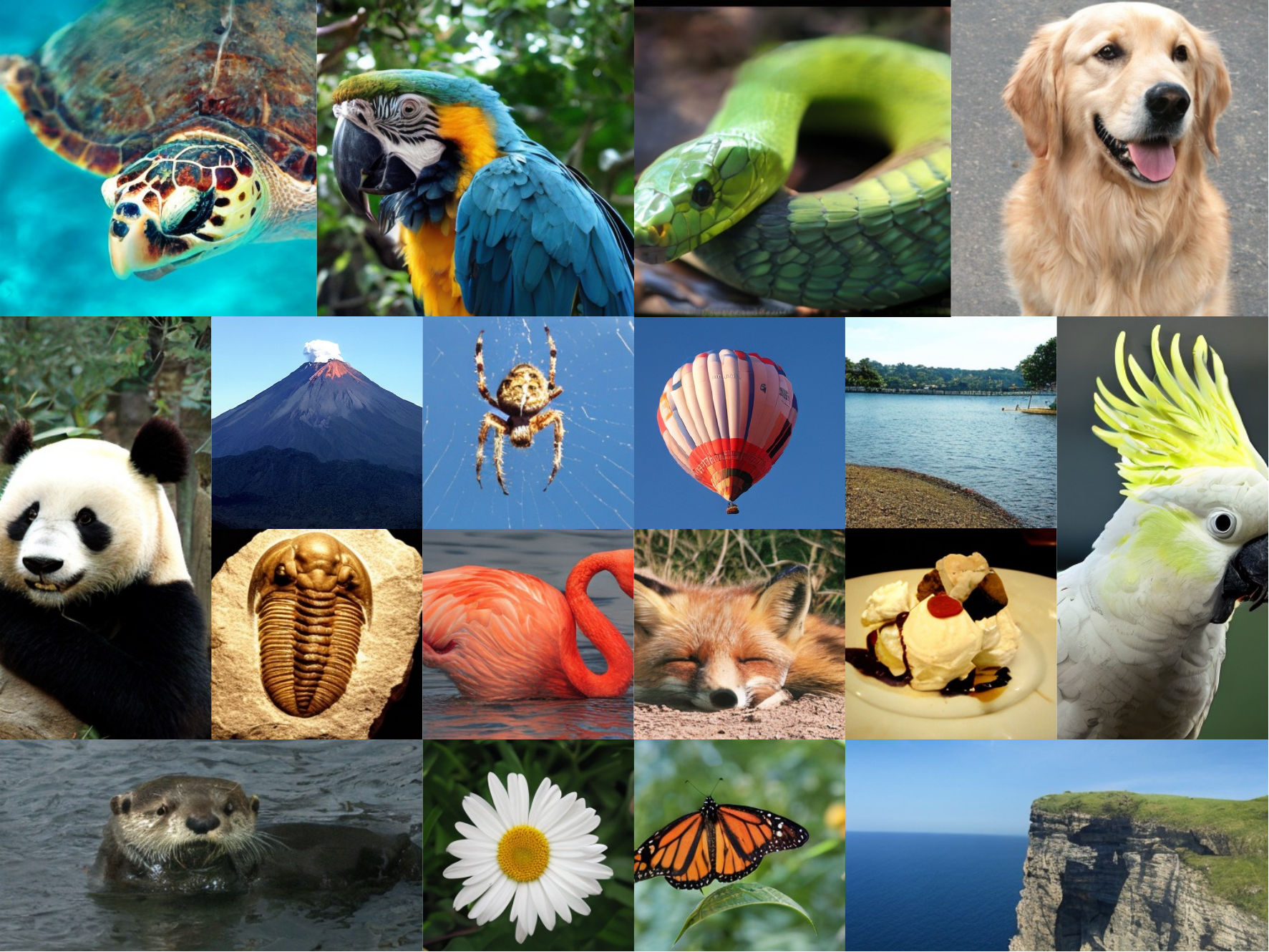}
    \caption{\textbf{Selected arbitrary-resolution samples (384x384, 224x448, 448x224, 256x256).} {\small Generated from a single FlowDCN-XL/2 model trained on ImageNet 256×256 resolution with CFG = 4.0.}}
    \vspace{-1.5mm}
    \label{fig:selected_samples}
\end{center}

\input{article/abs}
\input{article/introduction}
\input{article/related_works}
\input{article/method}
\input{article/experiment}

\input{article/conclusion}

{
\small
\bibliographystyle{unsrt}
\bibliography{nips24}
}

\newpage
\input{article/supp1}


\section*{Supplementary material}

## D. More Uncurated 56 × 56 Samples Visualization

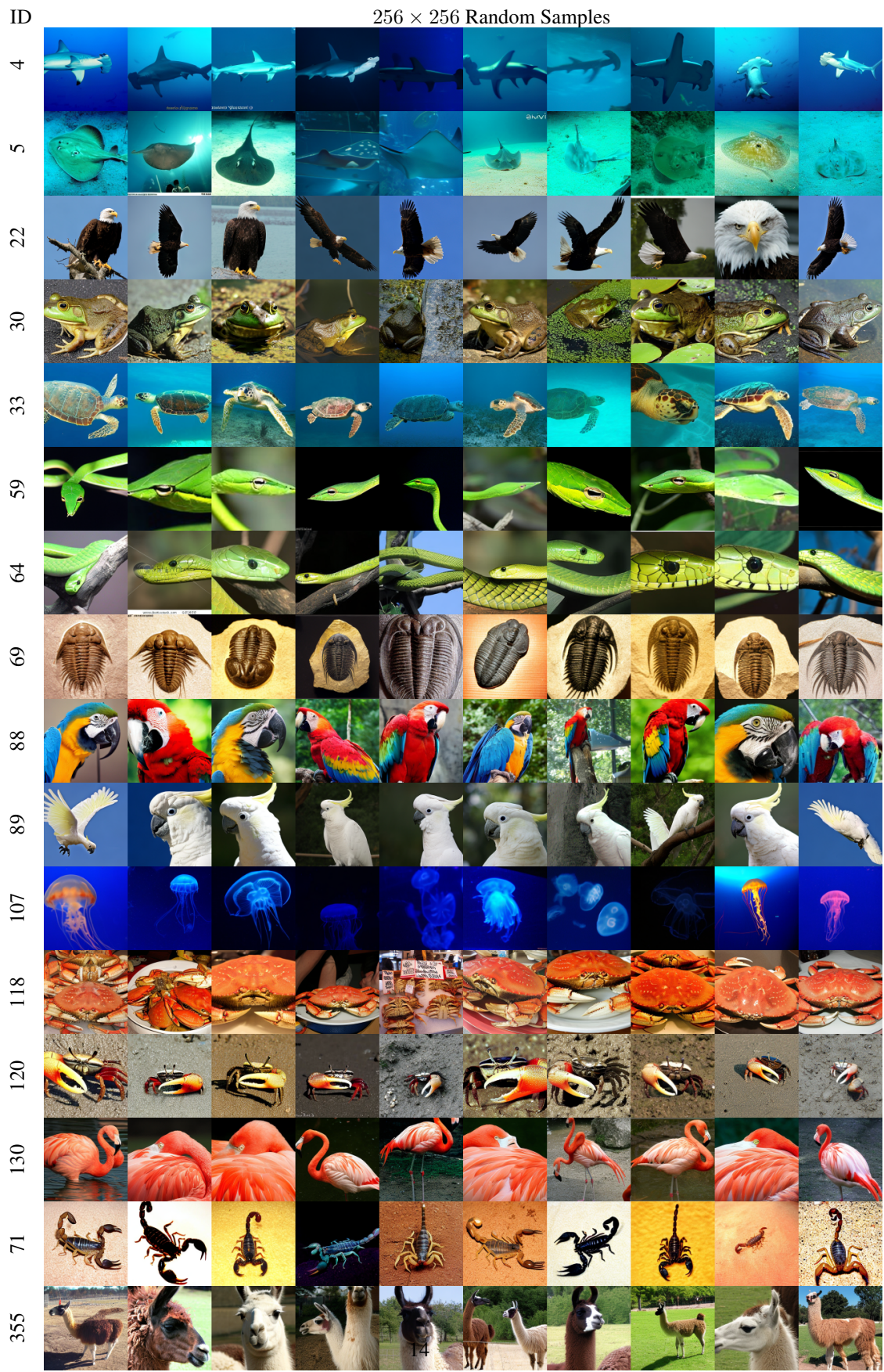

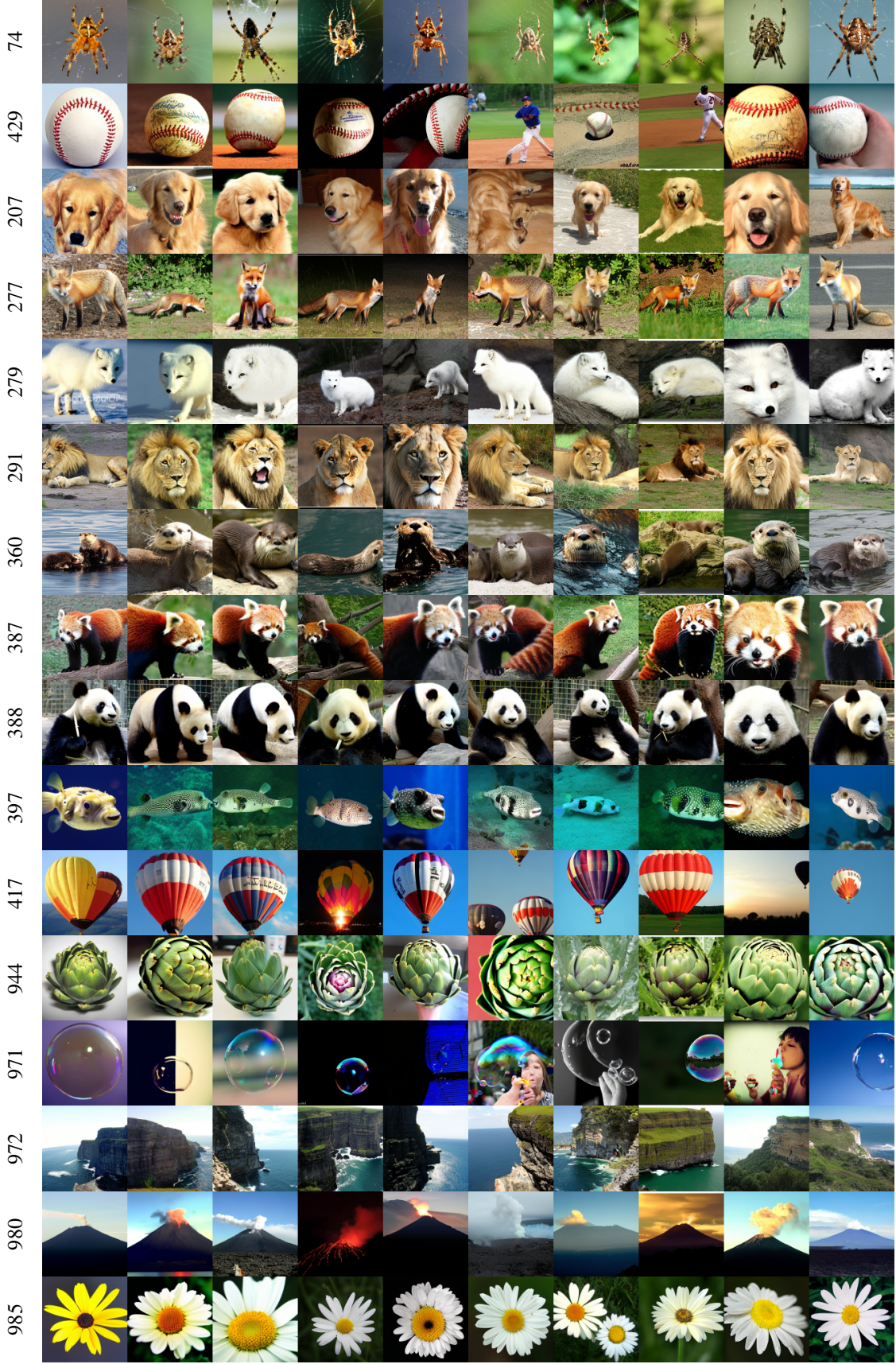

# E. Fewer Steps Uncurated ODE Samples $56 \times 56$ Compared to SiT

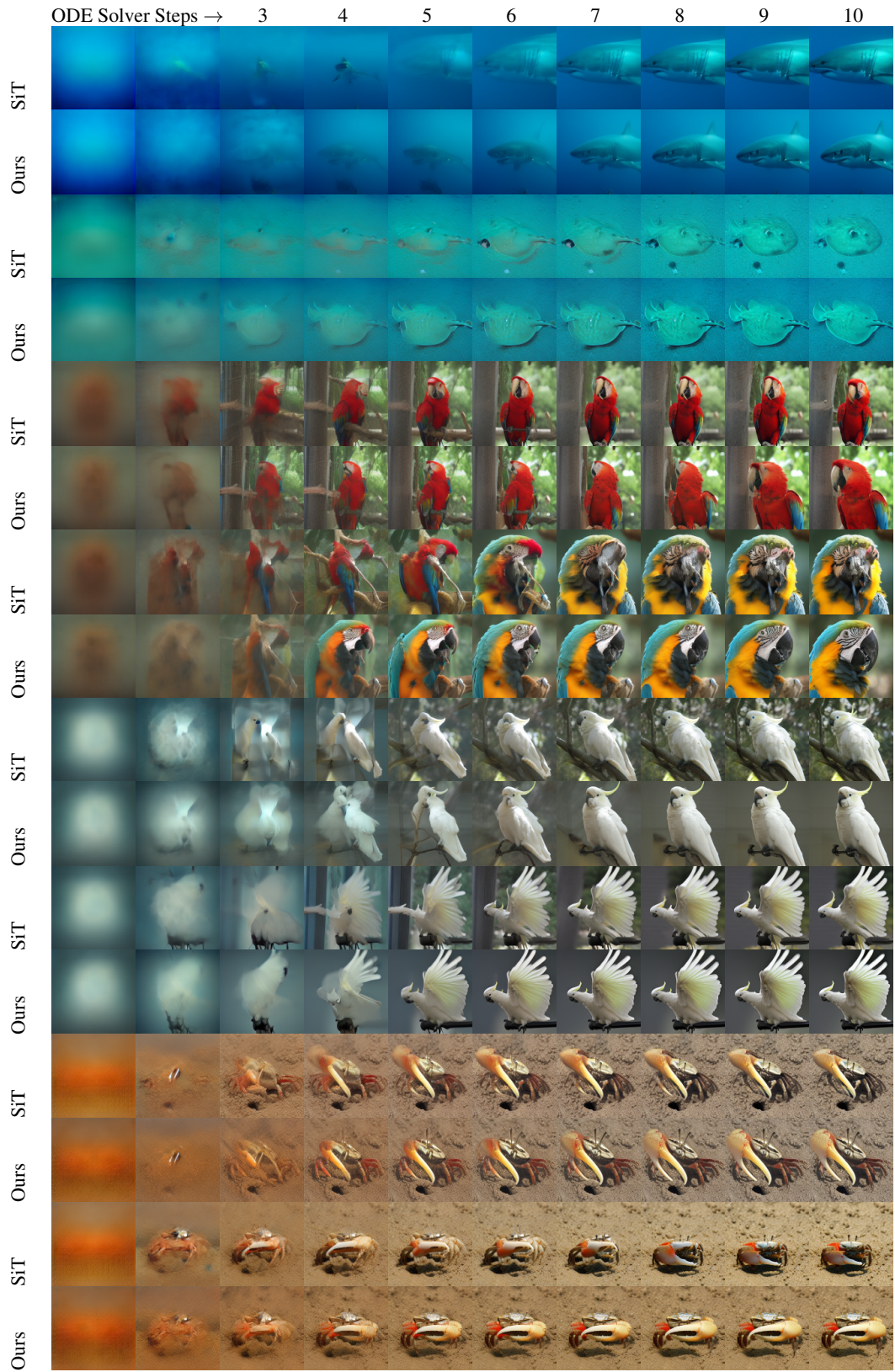

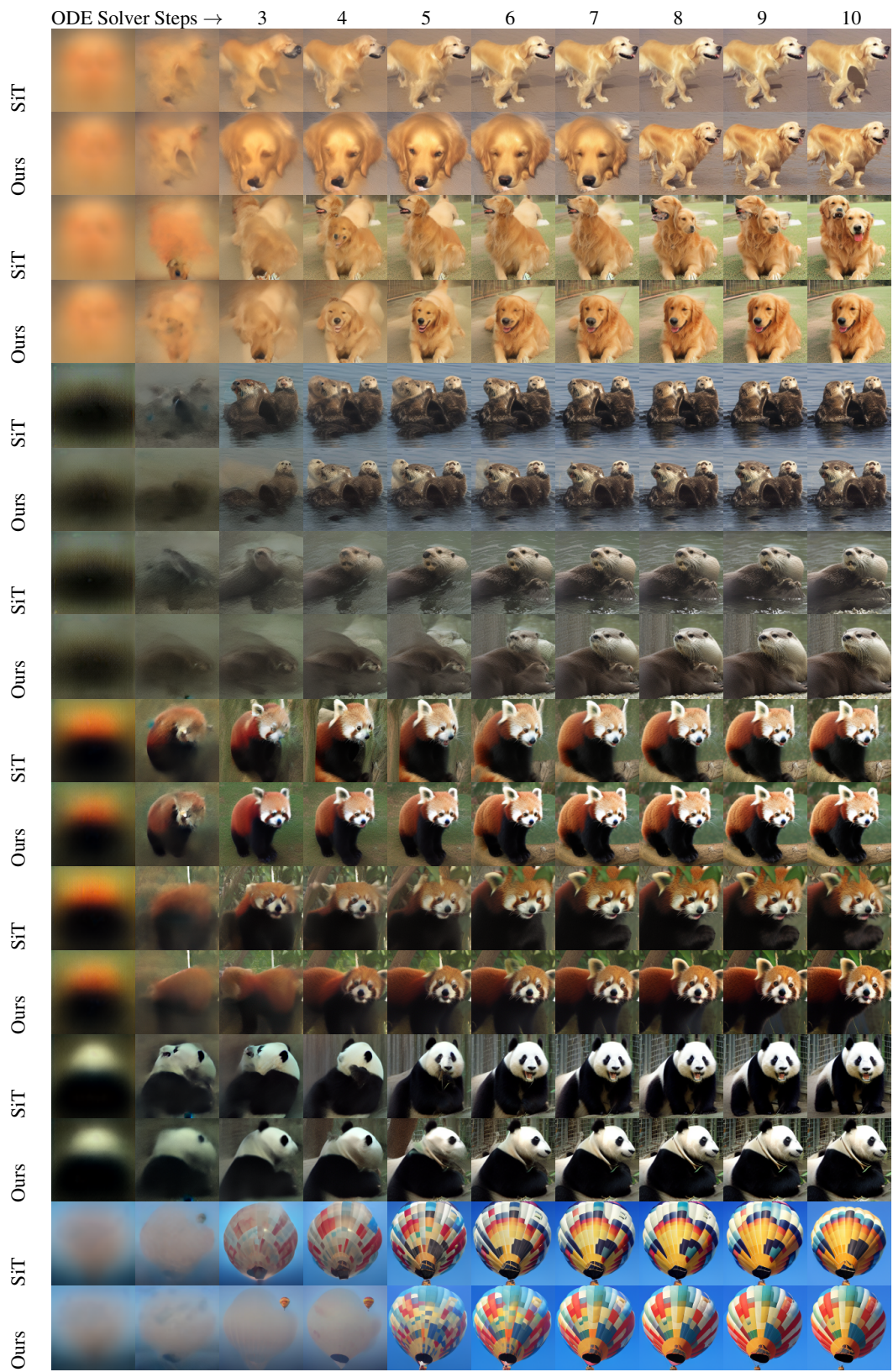
17

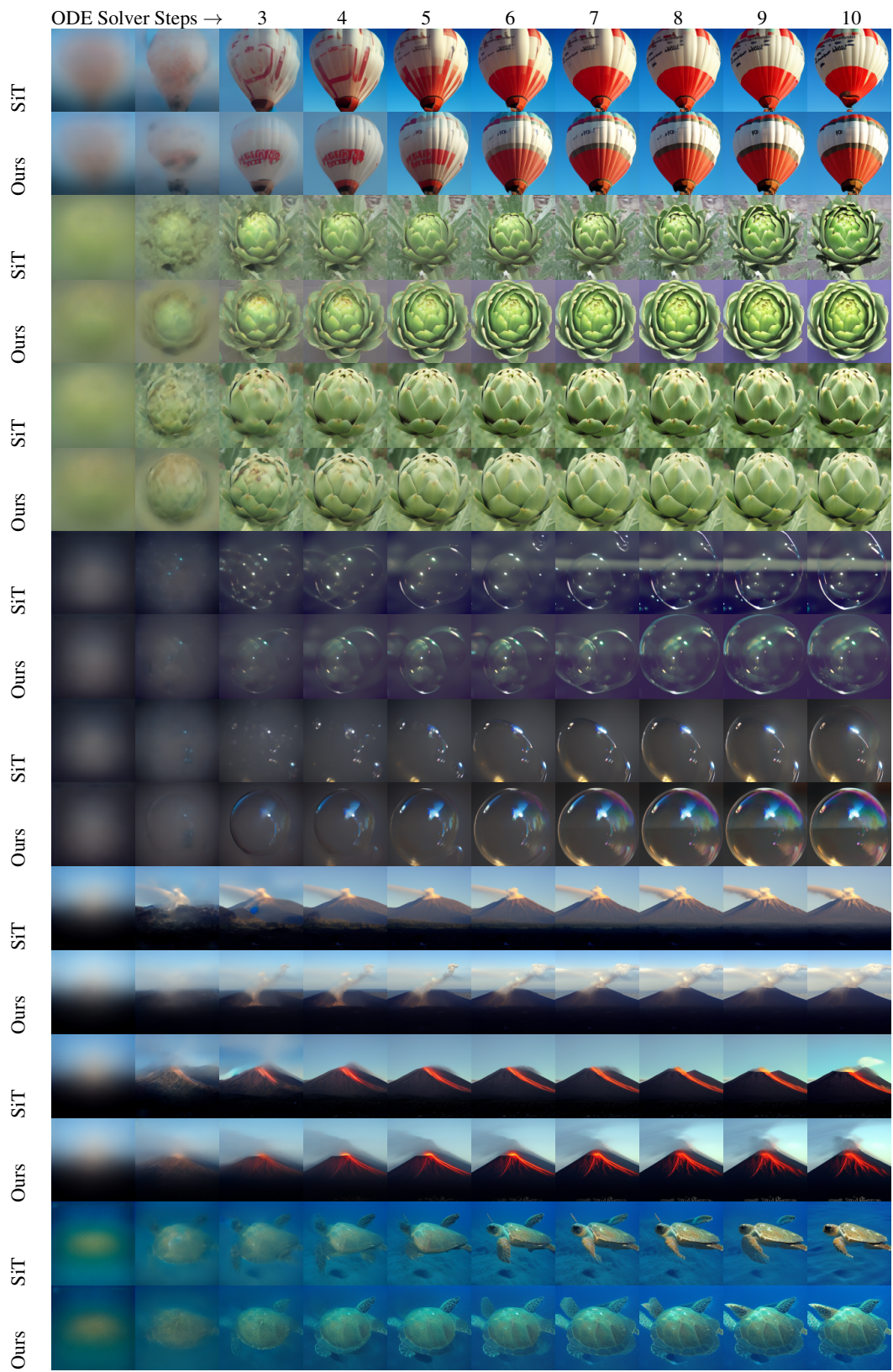

## F. Wide Images Samples (256 × 1024)

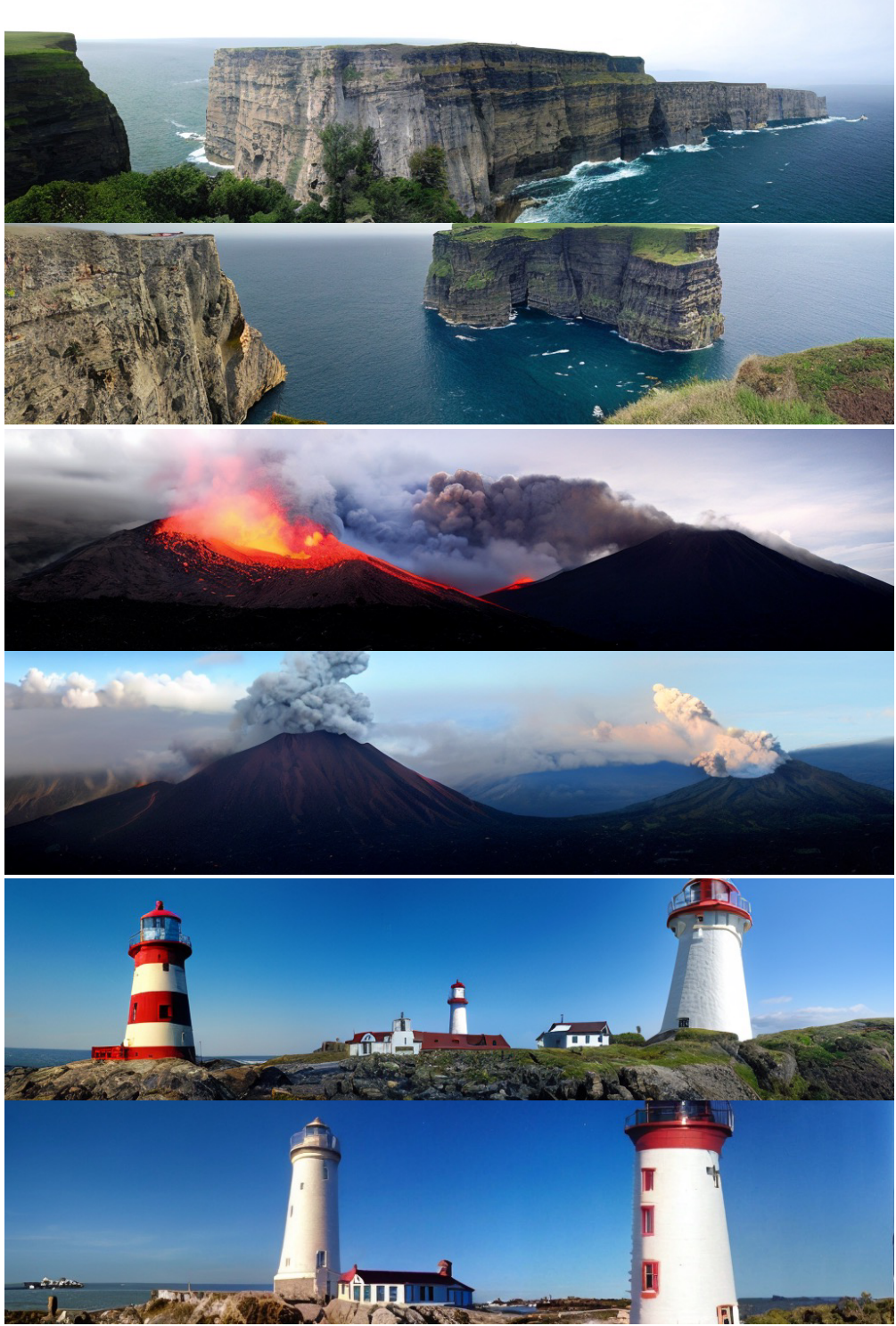

Figure 5: Wide images examples of Class ID (972 , 980, 437)

19

## G. More Uncarted Samples with $S_{max}$ Adjustment (512 → 256 and 256 → 512)

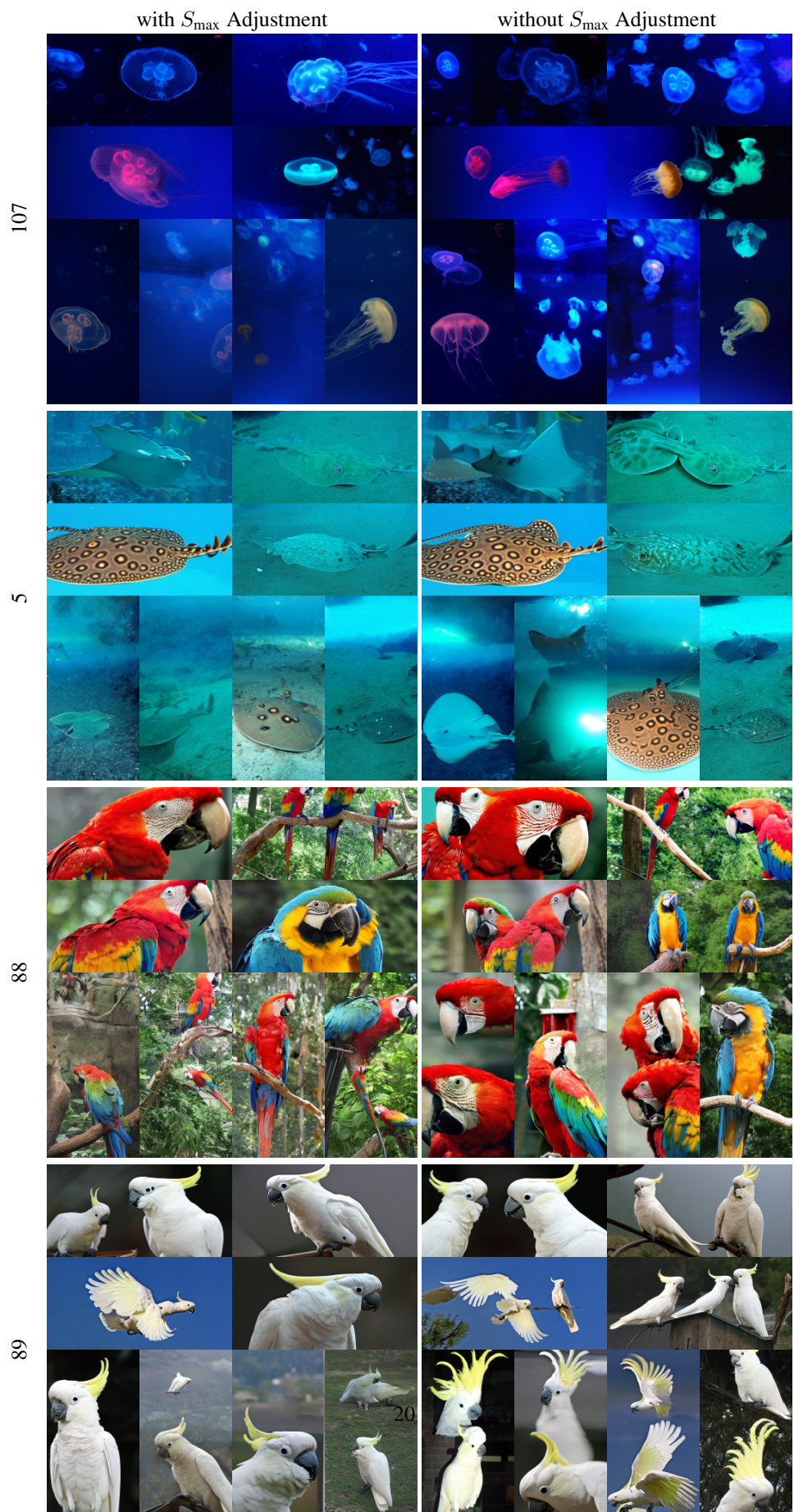

with $S_{max}$ Adjustment      without $S_{max}$ Adjustment

107

5

88

89

20

|  | with $S_{\max}$ Adjustment | without $S_{\max}$ Adjustment |
|---|---|---|

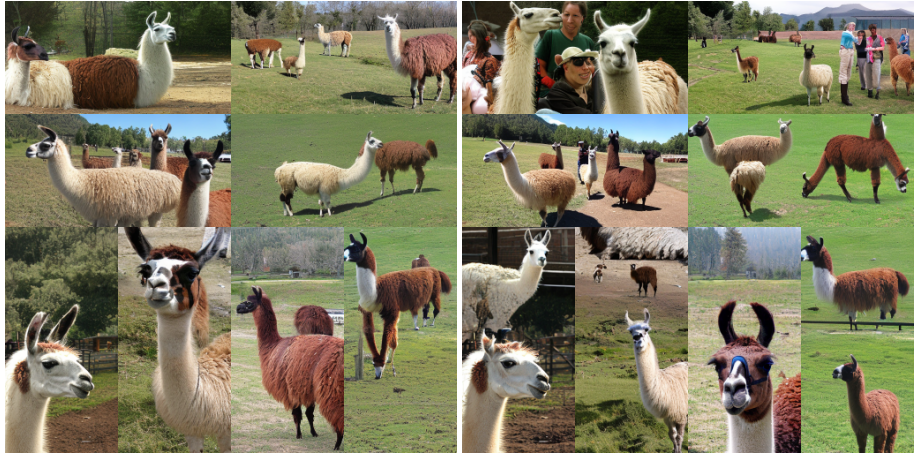

355

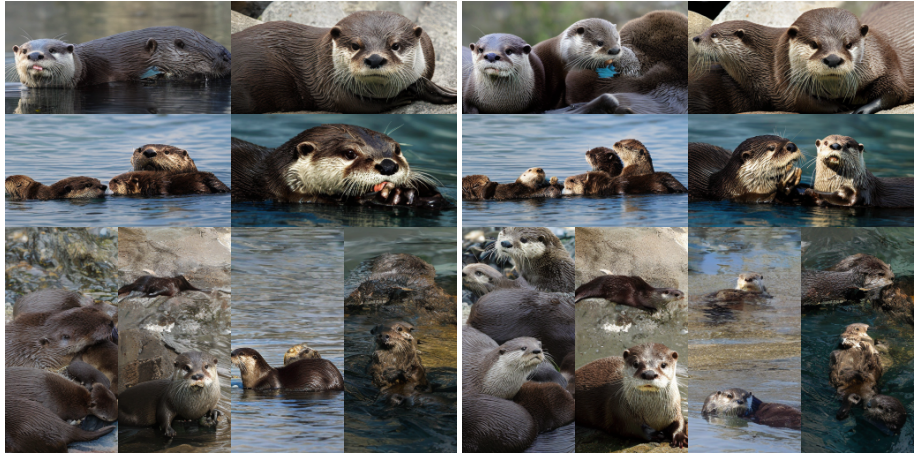

360

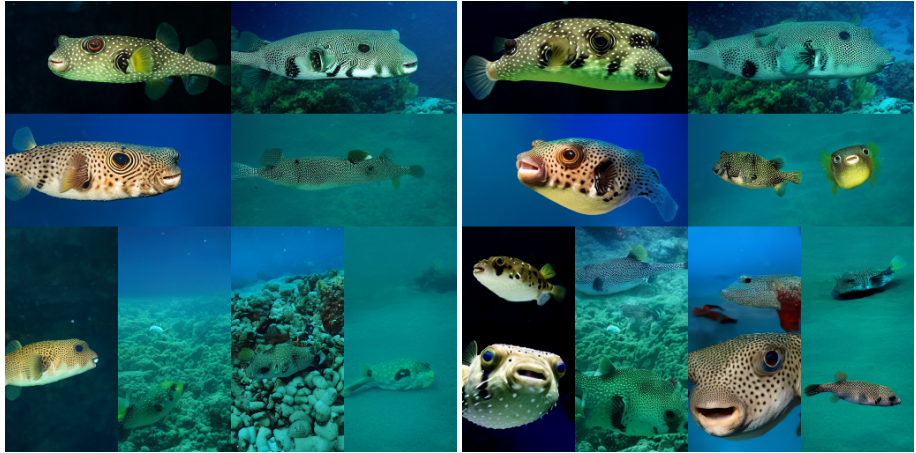

397

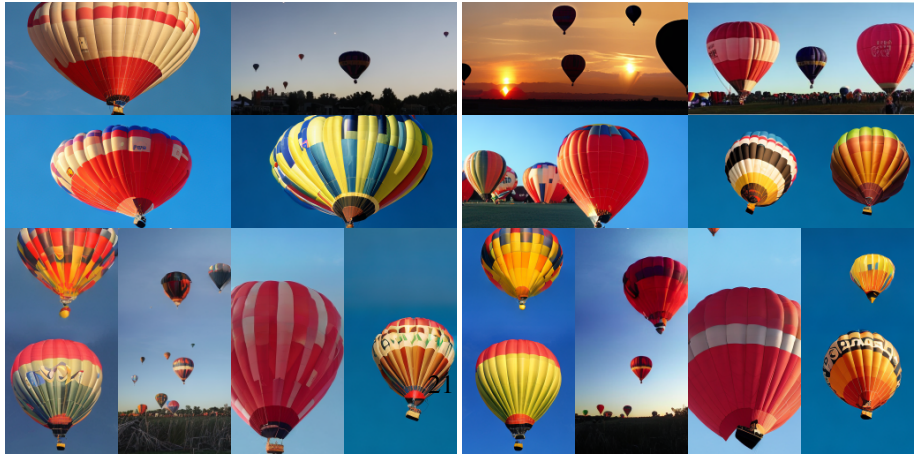

417

21

# NeurIPS Paper Checklist

1. **Claims**

   Question: Do the main claims made in the abstract and introduction accurately reflect the paper's contributions and scope?

   Answer: [Yes]

   Justification:

   Guidelines:
   - The answer NA means that the abstract and introduction do not include the claims made in the paper.
   - The abstract and/or introduction should clearly state the claims made, including the contributions made in the paper and important assumptions and limitations. A No or NA answer to this question will not be perceived well by the reviewers.
   - The claims made should match theoretical and experimental results, and reflect how much the results can be expected to generalize to other settings.
   - It is fine to include aspirational goals as motivation as long as it is clear that these goals are not attained by the paper.

2. **Limitations**

   Question: Does the paper discuss the limitations of the work performed by the authors?

   Answer: [Yes]

   Justification:

   Guidelines:
   - The answer NA means that the paper has no limitation while the answer No means that the paper has limitations, but those are not discussed in the paper.
   - The authors are encouraged to create a separate "Limitations" section in their paper.
   - The paper should point out any strong assumptions and how robust the results are to violations of these assumptions (e.g., independence assumptions, noiseless settings, model well-specification, asymptotic approximations only holding locally). The authors should reflect on how these assumptions might be violated in practice and what the implications would be.
   - The authors should reflect on the scope of the claims made, e.g., if the approach was only tested on a few datasets or with a few runs. In general, empirical results often depend on implicit assumptions, which should be articulated.
   - The authors should reflect on the factors that influence the performance of the approach. For example, a facial recognition algorithm may perform poorly when image resolution is low or images are taken in low lighting. Or a speech-to-text system might not be used reliably to provide closed captions for online lectures because it fails to handle technical jargon.
   - The authors should discuss the computational efficiency of the proposed algorithms and how they scale with dataset size.
   - If applicable, the authors should discuss possible limitations of their approach to address problems of privacy and fairness.
   - While the authors might fear that complete honesty about limitations might be used by reviewers as grounds for rejection, a worse outcome might be that reviewers discover limitations that aren't acknowledged in the paper. The authors should use their best judgment and recognize that individual actions in favor of transparency play an important role in developing norms that preserve the integrity of the community. Reviewers will be specifically instructed to not penalize honesty concerning limitations.

3. **Theory Assumptions and Proofs**

   Question: For each theoretical result, does the paper provide the full set of assumptions and a complete (and correct) proof?

   Answer: [NA]

Justification: not include theoretical results

Guidelines:
- The answer NA means that the paper does not include theoretical results.
- All the theorems, formulas, and proofs in the paper should be numbered and cross-referenced.
- All assumptions should be clearly stated or referenced in the statement of any theorems.
- The proofs can either appear in the main paper or the supplemental material, but if they appear in the supplemental material, the authors are encouraged to provide a short proof sketch to provide intuition.
- Inversely, any informal proof provided in the core of the paper should be complemented by formal proofs provided in appendix or supplemental material.
- Theorems and Lemmas that the proof relies upon should be properly referenced.

4. **Experimental Result Reproducibility**

   Question: Does the paper fully disclose all the information needed to reproduce the main experimental results of the paper to the extent that it affects the main claims and/or conclusions of the paper (regardless of whether the code and data are provided or not)?

   Answer: [Yes]

   Justification:

   Guidelines:
   - The answer NA means that the paper does not include experiments.
   - If the paper includes experiments, a No answer to this question will not be perceived well by the reviewers: Making the paper reproducible is important, regardless of whether the code and data are provided or not.
   - If the contribution is a dataset and/or model, the authors should describe the steps taken to make their results reproducible or verifiable.
   - Depending on the contribution, reproducibility can be accomplished in various ways. For example, if the contribution is a novel architecture, describing the architecture fully might suffice, or if the contribution is a specific model and empirical evaluation, it may be necessary to either make it possible for others to replicate the model with the same dataset, or provide access to the model. In general. releasing code and data is often one good way to accomplish this, but reproducibility can also be provided via detailed instructions for how to replicate the results, access to a hosted model (e.g., in the case of a large language model), releasing of a model checkpoint, or other means that are appropriate to the research performed.
   - While NeurIPS does not require releasing code, the conference does require all submissions to provide some reasonable avenue for reproducibility, which may depend on the nature of the contribution. For example
     (a) If the contribution is primarily a new algorithm, the paper should make it clear how to reproduce that algorithm.
     (b) If the contribution is primarily a new model architecture, the paper should describe the architecture clearly and fully.
     (c) If the contribution is a new model (e.g., a large language model), then there should either be a way to access this model for reproducing the results or a way to reproduce the model (e.g., with an open-source dataset or instructions for how to construct the dataset).
     (d) We recognize that reproducibility may be tricky in some cases, in which case authors are welcome to describe the particular way they provide for reproducibility. In the case of closed-source models, it may be that access to the model is limited in some way (e.g., to registered users), but it should be possible for other researchers to have some path to reproducing or verifying the results.

5. **Open access to data and code**

   Question: Does the paper provide open access to the data and code, with sufficient instructions to faithfully reproduce the main experimental results, as described in supplemental material?

Answer: [No]

Justification: We plan to opensource our code and implementation later.

Guidelines:

- The answer NA means that paper does not include experiments requiring code.
- Please see the NeurIPS code and data submission guidelines (https://nips.cc/public/guides/ odeSubmissionPolicy) for more details.
- While we encourage the release of code and data, we understand that this might not be possible, so "No" is an acceptable answer. Papers cannot be rejected simply for not including code, unless this is central to the contribution (e.g., for a new open-source benchmark).
- The instructions should contain the exact command and environment needed to run to reproduce the results. See the NeurIPS code and data submission guidelines (https://nips.cc/public/guides/ odeSubmissionPolicy) for more details.
- The authors should provide instructions on data access and preparation, including how to access the raw data, preprocessed data, intermediate data, and generated data, etc.
- The authors should provide scripts to reproduce all experimental results for the new proposed method and baselines. If only a subset of experiments are reproducible, they should state which ones are omitted from the script and why.
- At submission time, to preserve anonymity, the authors should release anonymized versions (if applicable).
- Providing as much information as possible in supplemental material (appended to the paper) is recommended, but including URLs to data and code is permitted.

6. **Experimental Setting/Details**

   Question: Does the paper specify all the training and test details (e.g., data splits, hyper-parameters, how they were chosen, type of optimizer, etc.) necessary to understand the results?

   Answer: [Yes]

   Justification:

   Guidelines:

   - The answer NA means that the paper does not include experiments.
   - The experimental setting should be presented in the core of the paper to a level of detail that is necessary to appreciate the results and make sense of them.
   - The full details can be provided either with the code, in appendix, or as supplemental material.

7. **Experiment Statistical Significance**

   Question: Does the paper report error bars suitably and correctly defined or other appropriate information about the statistical significance of the experiments?

   Answer: [No]

   Justification: Running experiments demands a lot of resources and time.

   Guidelines:

   - The answer NA means that the paper does not include experiments.
   - The authors should answer "Yes" if the results are accompanied by error bars, confidence intervals, or statistical significance tests, at least for the experiments that support the main claims of the paper.
   - The factors of variability that the error bars are capturing should be clearly stated (for example, train/test split, initialization, random drawing of some parameter, or overall run with given experimental conditions).
   - The method for calculating the error bars should be explained (closed form formula, call to a library function, bootstrap, etc.)
   - The assumptions made should be given (e.g., Normally distributed errors).
   - It should be clear whether the error bar is the standard deviation or the standard error of the mean.

- It is OK to report 1-sigma error bars, but one should state it. The authors should preferably report a 2-sigma error bar than state that they have a 96% CI, if the hypothesis of Normality of errors is not verified.
- For asymmetric distributions, the authors should be careful not to show in tables or figures symmetric error bars that would yield results that are out of range (e.g. negative error rates).
- If error bars are reported in tables or plots, The authors should explain in the text how they were calculated and reference the corresponding figures or tables in the text.

8. **Experiments Compute Resources**

    Question: For each experiment, does the paper provide sufficient information on the computer resources (type of compute workers, memory, time of execution) needed to reproduce the experiments?

    Answer: [Yes]

    Justification:

    Guidelines:
    - The answer NA means that the paper does not include experiments.
    - The paper should indicate the type of compute workers CPU or GPU, internal cluster, or cloud provider, including relevant memory and storage.
    - The paper should provide the amount of compute required for each of the individual experimental runs as well as estimate the total compute.
    - The paper should disclose whether the full research project required more compute than the experiments reported in the paper (e.g., preliminary or failed experiments that didn't make it into the paper).

9. **Code Of Ethics**

    Question: Does the research conducted in the paper conform, in every respect, with the NeurIPS Code of Ethics https://neurips.cc/public/EthicsGuidelines?

    Answer: [Yes]

    Justification:

    Guidelines:
    - The answer NA means that the authors have not reviewed the NeurIPS Code of Ethics.
    - If the authors answer No, they should explain the special circumstances that require a deviation from the Code of Ethics.
    - The authors should make sure to preserve anonymity (e.g., if there is a special consideration due to laws or regulations in their jurisdiction).

10. **Broader Impacts**

    Question: Does the paper discuss both potential positive societal impacts and negative societal impacts of the work performed?

    Answer: [Yes]

    Justification:

    Guidelines:
    - The answer NA means that there is no societal impact of the work performed.
    - If the authors answer NA or No, they should explain why their work has no societal impact or why the paper does not address societal impact.
    - Examples of negative societal impacts include potential malicious or unintended uses (e.g., disinformation, generating fake profiles, surveillance), fairness considerations (e.g., deployment of technologies that could make decisions that unfairly impact specific groups), privacy considerations, and security considerations.
    - The conference expects that many papers will be foundational research and not tied to particular applications, let alone deployments. However, if there is a direct path to any negative applications, the authors should point it out. For example, it is legitimate to point out that an improvement in the quality of generative models could be used to

generate deepfakes for disinformation. On the other hand, it is not needed to point out that a generic algorithm for optimizing neural networks could enable people to train models that generate Deepfakes faster.
- The authors should consider possible harms that could arise when the technology is being used as intended and functioning correctly, harms that could arise when the technology is being used as intended but gives incorrect results, and harms following from (intentional or unintentional) misuse of the technology.
- If there are negative societal impacts, the authors could also discuss possible mitigation strategies (e.g., gated release of models, providing defenses in addition to attacks, mechanisms for monitoring misuse, mechanisms to monitor how a system learns from feedback over time, improving the efficiency and accessibility of ML).

11. **Safeguards**

    Question: Does the paper describe safeguards that have been put in place for responsible release of data or models that have a high risk for misuse (e.g., pretrained language models, image generators, or scraped datasets)?

    Answer: [NA]

    Justification: Current models are only trained on small datasets.

    Guidelines:
    - The answer NA means that the paper poses no such risks.
    - Released models that have a high risk for misuse or dual-use should be released with necessary safeguards to allow for controlled use of the model, for example by requiring that users adhere to usage guidelines or restrictions to access the model or implementing safety filters.
    - Datasets that have been scraped from the Internet could pose safety risks. The authors should describe how they avoided releasing unsafe images.
    - We recognize that providing effective safeguards is challenging, and many papers do not require this, but we encourage authors to take this into account and make a best faith effort.

12. **Licenses for existing assets**

    Question: Are the creators or original owners of assets (e.g., code, data, models), used in the paper, properly credited and are the license and terms of use explicitly mentioned and properly respected?

    Answer: [Yes]

    Justification:

    Guidelines:
    - The answer NA means that the paper does not use existing assets.
    - The authors should cite the original paper that produced the code package or dataset.
    - The authors should state which version of the asset is used and, if possible, include a URL.
    - The name of the license (e.g., CC-BY 4.0) should be included for each asset.
    - For scraped data from a particular source (e.g., website), the copyright and terms of service of that source should be provided.
    - If assets are released, the license, copyright information, and terms of use in the package should be provided. For popular datasets, paperswithcode.com/datasets has curated licenses for some datasets. Their licensing guide can help determine the license of a dataset.
    - For existing datasets that are re-packaged, both the original license and the license of the derived asset (if it has changed) should be provided.
    - If this information is not available online, the authors are encouraged to reach out to the asset's creators.

13. **New Assets**

    Question: Are new assets introduced in the paper well documented and is the documentation provided alongside the assets?

Answer: [Yes]

Justification:

Guidelines:
- The answer NA means that the paper does not release new assets.
- Researchers should communicate the details of the dataset/code/model as part of their submissions via structured templates. This includes details about training, license, limitations, etc.
- The paper should discuss whether and how consent was obtained from people whose asset is used.
- At submission time, remember to anonymize your assets (if applicable). You can either create an anonymized URL or include an anonymized zip file.

14. **Crowdsourcing and Research with Human Subjects**

    Question: For crowdsourcing experiments and research with human subjects, does the paper include the full text of instructions given to participants and screenshots, if applicable, as well as details about compensation (if any)?

    Answer: [NA]

    Justification:

    Guidelines:
    - The answer NA means that the paper does not involve crowdsourcing nor research with human subjects.
    - Including this information in the supplemental material is fine, but if the main contribution of the paper involves human subjects, then as much detail as possible should be included in the main paper.
    - According to the NeurIPS Code of Ethics, workers involved in data collection, curation, or other labor should be paid at least the minimum wage in the country of the data collector.

15. **Institutional Review Board (IRB) pprovals or Equivalent for Research with Human Subjects**

    Question: Does the paper describe potential risks incurred by study participants, whether such risks were disclosed to the subjects, and whether Institutional Review Board (IRB) approvals (or an equivalent approval/review based on the requirements of your country or institution) were obtained?

    Answer: [NA]

    Justification:

    Guidelines:
    - The answer NA means that the paper does not involve crowdsourcing nor research with human subjects.
    - Depending on the country in which research is conducted, IRB approval (or equivalent) may be required for any human subjects research. If you obtained IRB approval, you should clearly state this in the paper.
    - We recognize that the procedures for this may vary significantly between institutions and locations, and we expect authors to adhere to the NeurIPS Code of Ethics and the guidelines for their institution.
    - For initial submissions, do not include any information that would break anonymity (if applicable), such as the institution conducting the review.



\end{document}

%% file: article/abs.tex
\begin{abstract}
\vspace{-2mm}
{ Arbitrary-resolution image generation still remains a challenging task in AIGC, as it requires handling varying resolutions and aspect ratios while maintaining high visual quality. Existing transformer-based diffusion methods suffer from quadratic computation cost and limited resolution extrapolation capabilities, making them less effective for this task. In this paper, we propose FlowDCN, a purely convolution-based generative model with linear time and memory complexity, that can efficiently generate high-quality images at arbitrary resolutions. Equipped with a new design of learnable group-wise deformable convolution block, our FlowDCN yields higher flexibility and capability to handle different resolutions with a single model.
FlowDCN achieves the state-of-the-art 4.30 sFID on $256\times256$ ImageNet Benchmark and comparable resolution extrapolation results, surpassing transformer-based counterparts in terms of convergence speed (only $\frac{1}{5}$ images), visual quality, parameters ($8\%$ reduction) and FLOPs ($20\%$ reduction). We believe FlowDCN offers a promising solution to scalable and flexible image synthesis.}
\end{abstract} 

%% file: article/introduction.tex
\section{Introduction}
\vspace{-0.5em}

Image generation is an important task in computer vision research, which is aimed at capturing the inherent data distribution of original image datasets and generating high-quality synthetic images through sampling. Diffusion models~\cite{ddpm, vp, edm, flow, flow2} have recently emerged as a highly promising foundation for training algorithms in image generation, outperforming GAN-based models~\cite{largegan, styleganxl} and Auto-Regressive models~\cite{maskgit} by a significant margin. The evolution of diffusion models is fast, transitioning from discrete forms~\cite{ddpm} to SDE-based continuous forms~\cite{vp, edm, flow, flow2, stochastic}. In a nutshell, diffusion models incrementally degrade an image through a time-dependent stochastic perturbation process and then learn the reverse process to restore the original image from its corrupted state.

Beyond theoretical advancements in diffusion models, the architecture of these models also significantly influences the quality of generated images. Many works~\cite{ddpm, adm, uvit} in the diffusion domain adopt a standard UNet architecture as the generation backbone, which consists of downsample blocks, upsample blocks, and long residual connections between these components. Inspired by the success of the vision transformer in perception tasks, DiT~\cite{dit} eliminates the long residual connection in favor of a pure transformer-based architecture. Through rigorous experiments, DiT demonstrates that the UNet inductive bias is not essential for achieving high performance in diffusion models~\cite{dit}. Meanwhile, PixArt~\cite{pixart, pixart_sigma} and SD3~\cite{sd3} venture further by significantly increasing the number of parameters, exploring new frontiers in model architecture and its impact on image generation.

When considering the generation of images at arbitrary resolution, diffusion transformers need to confront at least two primary challenges. The first is the \textit{quadratic computation cost}: the architecture of diffusion transformers employs attention mechanisms to aggregate spatial tokens. Owing to the dense nature of attention computations, high-resolution image generation inevitably leads to significant computation and memory demands, both scaling with $O(n^2)$ complexity. To address the quadratic computation challenge, some methods~\cite{diffusionssm} have adapted recurrent computational strategies from natural language processing. However, these adaptations do not fully capitalize on the strengths of autoregressive tasks and result in slower inference speeds due to the reduced parallelism inherent in RNN-based scanning. 
The second challenge is \textit{resolution extrapolation}: many diffusion transformers rely on absolute position embedding (APE)~\cite{attention} to incorporate positional information, introducing it at the onset of the model. This approach forces subsequent layers to become overfitted to the APE for providing positional context to the attention layers, which presents a significant barrier when extrapolating to different resolutions. To address this issue, FiT~\cite{fit} has turned to Rotary Positional Encoding~\cite{rope}, incorporating RoPE2D to enhance its resolution extrapolation capabilities. Nevertheless, FiT still requires a training pipeline tailored to arbitrary-resolution generation.

In contrast, convolutional models are the most common choice of visual encoders, boasting linear complexity and aggregating spatial features based on relative positions. With the support of modern convolution operators~\cite{dcnv4, convnext, replk}, convolutional models have demonstrated comparable performance or even surpassed transformers in perception tasks. This naturally leads us to inquire: \textit{Can modern convolutional networks achieve arbitrary-resolution generation efficiently and outperform transformer counterparts?} To answer this question, we opt for deformable convolution as the basic block for exploration in generation, owing to its superior performance in perception tasks.

Specifically, we propose a novel approach to decouple the scale and direction prediction of deformable convolution, giving rise to a group-wise multiscale deformable convolution block that enables efficient multiscale feature aggregation. By leveraging this block, we introduce FlowDCN, a modern purely convolution generative model that tackles arbitrary-resolution generation. Thanks to the new design of convolutional deformable block, our FlowDCN yields higher flexibility and capability to handle different resolutions with a single model. The experiments demonstrate that FlowDCN consistently surpasses its diffusion transformer counterparts, DiT~\cite{dit} and SiT~\cite{sit}. Notably, on the 256x256 ImageNet benchmark, FlowDCN achieves faster convergence, yielding SoTA sFid of 4.30 and FID of 2.13 under 1.5M steps with batch size 256, while exhibiting 20\% lower latency, 8\% fewer parameters, and approximately 20\% fewer floating-point operations (FLOPs). On the 512x512 ImageNet benchmark, FlowDCN achieves 4.53 sFid o and 2.44 FID under 100K finetuning steps with batch size 256.

Moreover, our FlowDCN offers a significant advantage in fast arbitrary-resolution generation, as it only requires linear time and memory complexity. Through visualization comparisons, our FlowDCN demonstrates substantially better visual quality even at extremely small sampling steps, such as 3, 4, and 5 steps. To further enhance its visual quality, we propose Scale Adjustment, a simpler technique for extrapolating resolution to unseen dimensions. Our results show that FlowDCN achieves comparable resolution extrapolation capabilities to highly tailored methods, underscoring its potential for generating high-quality images at various resolutions. The contributions can be summarized as:
\begin{itemize}
    \item { We decouple the scale and direction priors of deformable convolution and propose a Group-wise MultiScale Deformable Block. Building upon this block, we propose FlowDCN, a purely convolution-based generative model with high efficiency.}
    \item { On 256x256 ImageNet benchmark, under only 1.5M training steps, our FlowDCN-XL/2 achieves $2.13$ FID and SoTA $4.30$ sFID with Euler solver and classifier free guidance. }
    \item { On 512x512 ImageNet benchmark, under only 100K finetuning steps, our FlowDCN-XL/2 achieves $2.44$ FID and $4.53$ sFID with Euler solver and classifier free guidance. }
    \item { We propose a much simple and efficient resolution extrapolation method, deemed as Scale Adjustment. For arbitrary resolution generation, we achieve comparable results to highly tailored methods.}
\end{itemize}

%% file: article/related_works.tex

%% file: article/method.tex
\section{Preliminary}
\subsection{Linear-based Flow Matching.} Flow matching~\cite{flow, flow2} is a simple but powerful diffusion family. We incorporate linear-based flow matching as the training framework for its simplicity.  Given the image sampled $x$ from training distributions and the noise $\epsilon$ sampled from a Gaussian distribution, linear-based flow matching forward process interpolate $x_t$ with $x$ and $\epsilon$ using the following equation:
\begin{equation}
\label{eq:xt_linear}
    x_t = tx + (1-t)\epsilon.
\end{equation}
The velocity field of linear-based flow matching~\cite{flow, flow2} is defined as \cref{eq:vt_linear}. We train our FlowDCN to predict the time-dependent velocity field between $x$ and $\epsilon$:
\begin{equation}
    v_t(x_t) = x - \epsilon. \label{eq:vt_linear}
\end{equation}
During training, the flow matching objective directly regresses the target velocity:
\begin{equation}
    \mathcal{L}_{v} = \int_0^1 \mathbb{E}[\parallel v_{\theta}(x_t, t) - v_t(x_t) \parallel^2]dt.
\label{eq:loss}
\end{equation}
For sampling, the common ODE/SDE solver \eg. Euler method, Heun method can be employed.  

\subsection{Deformable Convolution Revisited}
Given an image feature $\mathbf{x} \in \mathbb{R}^{H\times W \times D}$, deformable convolution predicts the deformable field $\Delta \mathbf{P}(\textbf{x}) \in \mathbb{R}^{H\times W \times G \times K\times2}$ and the dynamic weights $\mathbf{W}(\mathbf{x}) \in \mathbb{R}^{H\times W \times G\times K}$ from the image feature $\mathbf{x}$. Specifically, $H$ and $W$ represent the height and width of the feature spatial shape, $D$ is the feature channel, $K$ is the number of sampling points, and $G$ is the number of groups in the deformable convolution operation. The deformable field and dynamic weights are computed as \cref{eq:dcn_weight_deformables}:
\vspace{-1mm}
\begin{align}
    \label{eq:dcn_weight_deformables}
    \Delta \mathbf{P}(\mathbf{x}) &= \mathbf{W}_{\text{deformable}}^T \mathbf{x} + \mathbf{b}_{\text{deformable}}, \\
    \mathbf{W}(\mathbf{x}) &= \mathbf{W}_{\text{weight}}^T \mathbf{x} + \mathbf{b}_{\text{weight}}. 
\end{align}
For a specific group $g$ in deformable convolution, the sampling position is determined by the base feature position $p_0$, sampling position prior $p_k$, and predicted deformable $\Delta p_k$ from $\Delta \mathbf{P}(\mathbf{x})$ for the $k$-th sampling point. The dynamic weight $w_k$ is provided from $\mathbf{W}(\mathbf{x})$. The deformable convolution aggregates $K$ sparse spatial features according to the sampling location and dynamic weight as following:
\vspace{-3mm}
\begin{align}
\label{eq:dcn}
    \mathbf{y}^g(p_0) &= \sum_{k=0}^K w^g_k \mathbf{x}^g(p_0 + p_k + \Delta p_k(\mathbf{x})),  \\
    \mathbf{y}   &= \text{concat}(\mathbf{y}^1, ~\mathbf{y}^2, ~...., ~\mathbf{y}^{G}).
\end{align}
The predefined spatial position prior $p_k$ is initialized from the regular convolution, commonly using ${(-1, -1), (-1, 0),... (0,0),...(1, 1)}$ as the predefined value.

Deformable convolution introduces long-range dependencies and dynamic aggregation into regular convolutions, bridging the gap between convolution and multi-head self-attention~\cite{dcnv3}. Thus, deformable convolution shares the efficiency merit of convolution and the dynamics merit of the attention mechanism. In most scenarios, DCN-like architectures are more powerful than common CNNs, we provide comparison experiments of DCN and CNN of flow matching training. Notably, deformable convolution directly predicts the dynamic weights and only aggregates limited features from spatial locations, enjoying a relatively sparse computation diagram. A deformable convolution operator only requires {\small $\frac{4KHWC}{G}$} FLOPs for computation when employing bilinear sampling to aggregate features.

\section{Method}
\subsection{MultiScale Deformable Convolution}
\label{sec:msdcn}
The original deformable convolution has been widely adopted in hierarchical model architectures~\cite{dcnv2, dcnv3, dcnv4} for perception tasks. However, these models typically progressively downsample the feature maps to increase the reception field growth rate. In contrast, image generation tasks require outputs with more high-frequency details and low-level information. From this perspective, progressively downsampling features would lead to the loss of high-frequency details. One possible solution is to introduce long residual connections to generation models~\cite{uvit, adm}. However, in practice, this approach demands caching image features from the encoder part, which increases peak memory usage during model inference.

To strike a balance between receptive fields and high-frequency details, we propose decoupling the deformable field into scale and direction, and introduce a novel multiscale deformable convolution. Unlike previous deformable convolutions, our approach assigns different scale priors to different groups. 

\paragraph{Decoupling deformable field to direction and scale.} The original deformable convolution directly regresses the deformable field to learn an unbounded and adaptive sampling point generator. However, the vast image spatial range poses a challenge to the learning process, as it leads to unstable regression of the deformable range when extrapolating from local neighbors to distant feature locations. We tackle this problem by decoupling the direction and scale of the deformable field. Specifically, we reorganize the sampling point formulation in \cref{eq:decouple_ds}. 
\begin{equation}
    s(\mathbf{x}) = S_\text{max}*\text{sigmoid}(\mathbf{W}_s^T \mathbf{x}),
\end{equation}
\begin{equation}
    \label{eq:decouple_ds}
    p = p_0 + s(\mathbf{x})*(p_k + \Delta p_k(\mathbf{x})),
\end{equation}
where $s(\mathbf{x})$ is the learnable scale predicted from the image feature $\mathbf{x}$. $S_\text{max}$ is the max scale value of the given deformable convolution, we leave it as a hyper-parameter only related to input resolution, thus we can manually tune it according to input resolution, details are placed in \cref{sec:abs_sampling}.

\paragraph{Group-wise multiscale deformable convolution.} To keep high-resolution feature maps and own a large reception field growth rate, we propose to assign different scale priors to different groups. This allows deformable groups with large scale priors to aggregate long-dependency features, while those with small scale priors aggregate short-dependency features as follows:
\begin{equation}
    \label{eq:dcn_scale}
    s^g(\mathbf{x}) = S_\text{max}*\text{sigmoid}(\mathbf{W}_s^T \mathbf{x} + s_0^g),
\end{equation}
\begin{equation}
    \label{eq:decouple_ds2}
    p = p_0 + s^g(\mathbf{x})*(p_k + \Delta p_k(\mathbf{x})).
\end{equation}

Specifically, we initialize the scale priors with \cref{eq:init_s0} and initialize $\mathbf{W}_s$ with zeros to obtain linearly increased $\text{sigmoid}(s_0^g)$ along group axis:
\begin{equation}
    \label{eq:init_s0}
    s_0^{g+1} = \log({\frac{g}{G-g}}).
\end{equation}

\subsection{Flow-based Deformable Convolution Generative Model}
We introduce our novel diffusion generation architecture, dubbed FlowDCN. Rather than directly adopting tailored architectures for image generation, such as long residuals and normalization techniques, we aim to explore the generative capabilities of deformable convolution-based architectures in a faithful manner. To this end, we deliberately discard long residual connections and opt to build a pure convolution-based generative model, preserving the unique characteristics of DCN-like models as much as possible. For training and sampling, we leverage the powerful flow matching algorithm to align our model with the state-of-the-art SiT~\cite{sit}.

\paragraph{Deformable convolution generative model.}
\begin{figure}
    \centering
    \subfloat[\textbf{FlowDCN Architecture.} {\small Our FlowDCN consists of stacked MultiScaleDCN blocks and SwiGLU blocks. We also employ RMSNorm to stabilize training.}\label{fig:flowdcn}]{
    \includegraphics[width=0.4\linewidth]{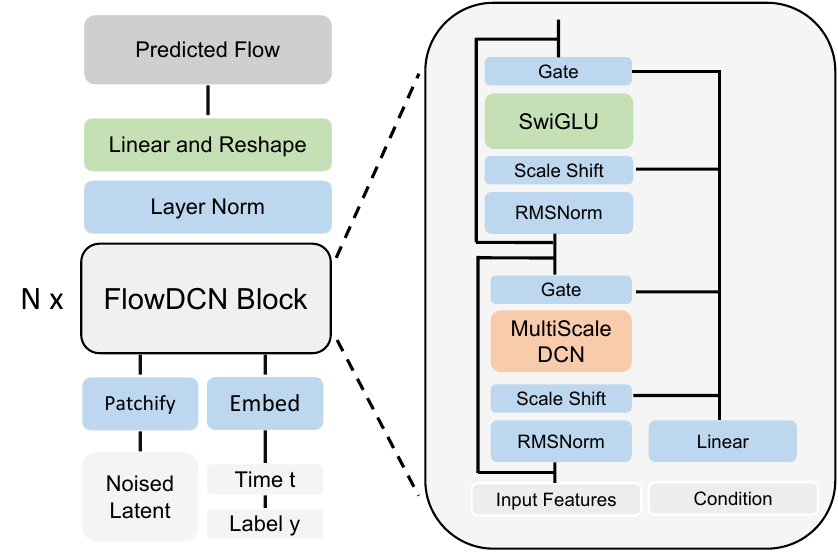}
    }\hspace{0.5em}
    \subfloat[\textbf{MultiScale DCN Block.} {\small Dynamic weight and scale\& direction deformable field are predicted from input features, then merged with priors to form the deformable kernels to extract features.} \label{fig:msdcn}]{
    \includegraphics[width=0.5\linewidth]{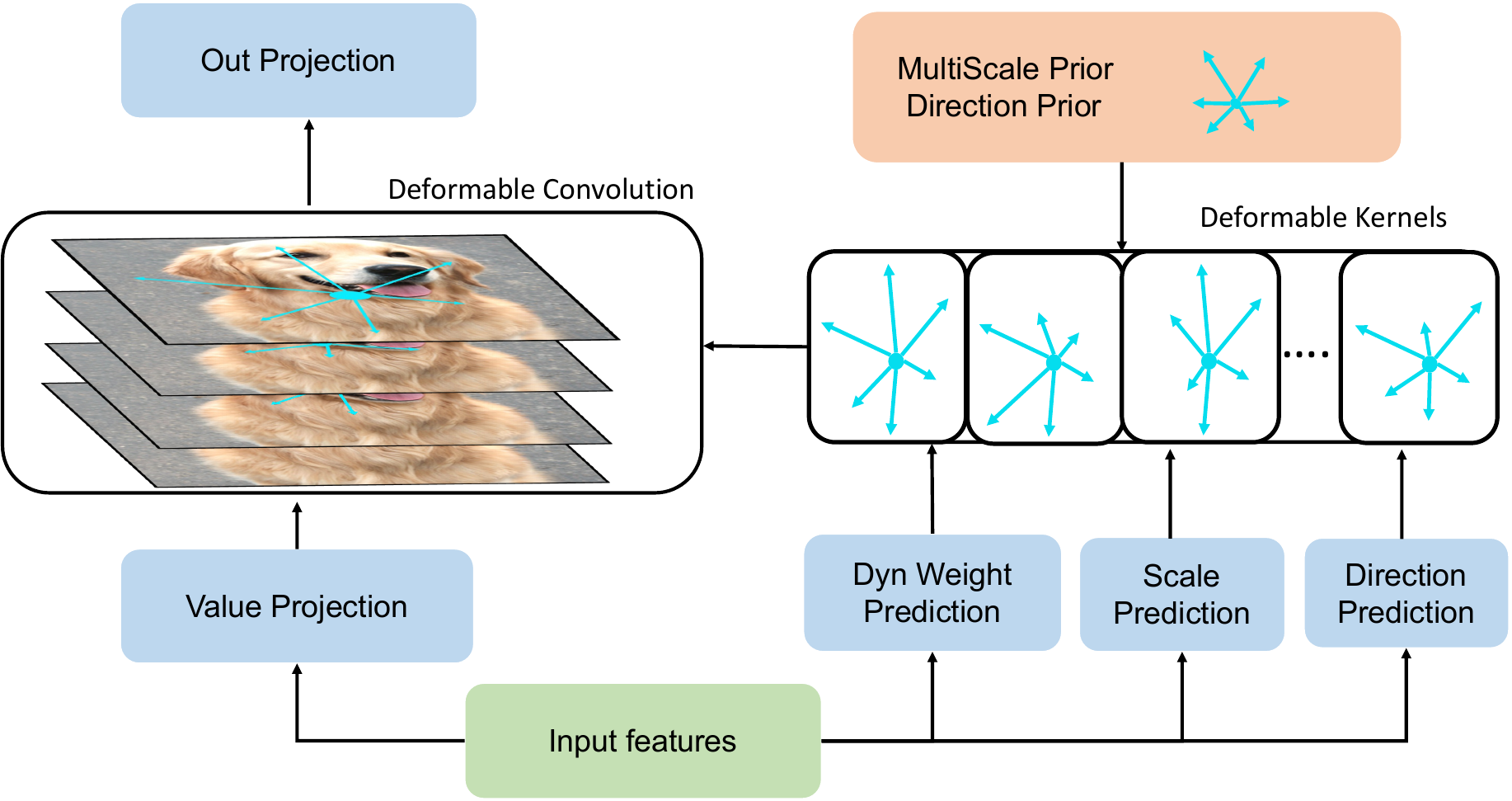}
    }
    \caption{ \textbf{The Architecture of Our FlowDCN and MultiScale DCN Block.}}
    \vspace{-1em}
\end{figure}
The model architecture is illustrated in \cref{fig:flowdcn}. We aim to build a pure DCN-like generative model to explore the generation ability of DCN-like~\cite{dcnv4} architectures. To match the base resolution of model input with DiT~\cite{dit} and SiT~\cite{sit}, we similarly patchify the noisy input via convolution.
Inspired by DiT~\cite{dit}, we inject the timestep and label conditions through adaLN-Zero~\cite{dit, adain}. 
The basic block is formulated as \cref{eq:block}. Drawing inspiration from LLaMA~\cite{llama1, llama2}, we replace vanilla FFN and LayerNorm with SwiGLU and RMSNorm, respectively. Note we also provide FFN and LayerNorm version FlowDCN for fair comparisons:
\begin{align}
    \label{eq:block}
    \mathbf{x}_1 &= \mathbf{x} + \text{AdaLN}(\mathbf{y}, \mathbf{t}, \text{MultiScale-DCN}(\mathbf{x})),\\
    \mathbf{x}_2 &= \mathbf{x}_1 + \text{AdaLN}(\mathbf{y}, \mathbf{t}, \ \ \text{SwiGLU}(\mathbf{x}_1)).
\end{align}

\subsection{Arbitrary Resolution Sampling}
\label{sec:abs_sampling}

We denote the training resolution as $H_\text{train} \times W_\text{train}$ and the inference resolution as $H_\text{test} \times W_\text{test}$. Notably, our FlowDCN is capable of handling arbitrary resolution that differs from the training resolution. As a reminder, the multiscale deformable convolution block aggregates features based on predicted scales and directions according to Equation (\cref{eq:decouple_ds}). In practice, the predicted scale of the multiscale deformable convolution layer is typically fitted to match the training resolution distribution. However, this limits the reception fields of the image features when encountering unseen resolution, ultimately hurting the global semantic consistency~\cite{scalecrafter, hidiffusion}. To improve the global semantic consistency, we propose adjusting the scaling factor based on the relative ratio between the training resolution and inference resolution. 

\paragraph{Adjust $S_\text{max}$ to match inference resolution.}  As shown in \cref{eq:dcn_scale}, $S_\text{max}$ controls the maximum sampling range in multiscale deformable convolution. As discussed in \cref{sec:msdcn}, we treat it as a resolution-dependent hyperparameter.  It is straightforward to observe that scaling $S_\text{max}$ with the relative aspect ratio between train size and inference size could match the reception field between train and inference:
\begin{align}
    s^g_h(\mathbf{x}) &= \text{sigmoid}(\mathbf{W}_s^T \mathbf{x} + s_0^g)\cdot S_\text{max} \cdot \frac{H_\text{test}}{H_\text{train}}, \\
    s^g_w(\mathbf{x}) &= \text{sigmoid}(\mathbf{W}_s^T \mathbf{x} + s_0^g)\cdot S_\text{max} \cdot \frac{W_\text{test}}{W_\text{train}}.
    \vspace{-1em}
\end{align}

%% file: article/experiment.tex
\section{Experiments}
We conduct experiments on 32x32 CIFAR10 and 256x256 ImageNet datasets. The training batch size is set to 256. Similar to SiT~\cite{sit} and DiT~\cite{dit}, we use Adam optimizer~\cite{adam} with a constant learning rate 0.0001 during the whole training. We do not adopt any gradient clip techniques for fair comparison.
For 32x32 CIFAR10 dataset, we train our model for 25000 steps. As for 256x256 ImageNet dataset, we train for 1.5M steps. We use $8\times$A100 GPUs as the default training hardware.

\paragraph{Efficient deformable convolution implementation.} 
Although DCNv4~\cite{dcnv4} proposes a much faster deformable convolution implementation, it is not tailored for image generation input shape. For resolution below $512\times512$, there are fewer spatial tokens (only $16\times16$ tokens for $256\times256$ resolution) to fully utilize sparse computation strengths, thus DCNv4 exhibits even worse latency compared to attention. To remedy high latency of deformable convolution for low-resolution scenery, we decide to leverage shared memory to reduce the latency of random sampling in deformable convolutions, deemed as DeformConv(shm).  We place the performance benchmark at \cref{tab:op_low_res}. For high-resolution scenery, We also re-implement DeformConv(DCNv4) in Triton-lang as DeformConv(Triton-lang) to leverage the strengths of compiler~\cite{triton, llvm} to find suitable hyperparameters. 

\input{article/tables/op_speed}

\subsection{32x32 CIFAR Dataset} 
The CIFAR10 dataset\cite{cifar}, comprising 50,000 32x32 small-resolution images from 10 distinct class categories, is considered an ideal benchmark to validate the design of our MultiScale deformable block due to its relatively small scale. We select SiT-S/2 as a comparison baseline, as it also leverages the flow-matching framework. For sampling, we employ the Euler stochastic solver with 1000 sampling steps to generate images. We report the FID~\cite{fid}, sFID~\cite{sfid}, and Inception Score~\cite{is} as the primary metrics to evaluate the performance of our model.

\input{article/tables/cifars}
\paragraph{Compare with baseline SiT.} We summarise the metrics of our FlowDCN and SiT in \cref{tab:compare_cifar10}. Our FlowDCN achieves 5.47 fid, surpassing its counterpart SiT with 2.0 fid margins. Additionally, our model performs slightly better in terms of sFID and Inception Scores, further demonstrating its superiority.

\paragraph{Group-wise multiscale design.} As showed in \cref{tab:compare_cifar10}, we denote the variant of FlowDCN that uses vanilla deformable convolution instead of Multiscale deformable convolution as \textit{w/o MultiScale}. Notably, the absence of group-wise multiscale deformable convolution leads to a 0.25 FID performance degradation. This result demonstrates the effectiveness and power of our proposed group-wise multiscale mechanism.

\paragraph{Prior initialization.} By default, we manually initialize the direction priors with predefined grids $\{(-1, -1), (-1, 0), ... (0,0), ...(1, 1)\}$, and initialize the scale priors with linearly increased scale along group axis. We also experiment with randomly initialized direction and scale priors in \cref{tab:compare_cifar10}, donated as \textit{w/o PriorInit}. Random initialization shows slight performance degradation.

\paragraph{Sampling points.} In \cref{tab:kernel_size}, We train FlowDCN with varying kernel sizes $K$ and observe that the performance consistently improves as the number increases. Specifically, using 32 points to aggregate features, FlowDCN achieves a FID score of 5.13 and an sFID score of 4.43. However, to maintain a relatively sparse pattern, we choose $K=9$ as the default setting, striking a balance between performance and computational efficiency.

\paragraph{Fixed direction priors.} In \cref{tab:priors}, we present the results of training FlowDCN with different prior learning settings. Notably, we find that the fixed direction prior $p_k$ in \cref{eq:decouple_ds} achieves better results compared to the learnable direction prior. We hypothesize that the learnable direction prior may cause the learning of the deformable field to become unstable, leading to inferior performance. 

\paragraph{Learnable relative scale.} In the ${s}(\mathbf{x})$ column of \cref{tab:priors}, the notation "\textit{learn}" indicates that we predict a relative scale of the deformable fields in addition to the learnable scale priors $s_0^g$ ~( $\mathbf{W}_s^T\mathbf{x}$ in \cref{eq:dcn_scale}), whereas "\textit{fixed}" does not predict the relative scales ${s}(\mathbf{x})$ in the deformable field. Learning a relative scale for each feature in \cref{tab:priors} achieves better results of 5.47 FID.

\subsection{256$\times$256 ImageNet Dataset}
Based on our analysis, we select the MultiScale deformable convolution with a kernel size of $K=9$ as the basic block for our Imagenet experiments. Our default setting involves fixing the direction priors and learning relative scales from the deformable field. We manually initialize the direction priors with predefined grids and initialize the scale priors with linearly increased scales. To generate images, we employ an Euler-Maruyama solver with 250 steps for stochastic sampling. We report the FID, sFID, Inception Score, and Precision \& Recall as the primary metrics to evaluate the performance of our model. 
\input{article/tables/imagenet256}

\paragraph{Metrics comparison with baseline SiT.}
We present the performances of different-size models at 400K training steps in \cref{tab:sit_sde}. From Small to XL-size models, our FlowDCN model family consistently outperforms its counterpart DiT~\cite{dit} and SiT~\cite{sit} with significant margins. Without RMS/SwiGLU, our FlowDCN-B/2 degrades with 0.6 FID gains but still surpasses SiT by a large margin. In addition to its superior performance and convergence speed, our FlowDCN also boasts a remarkable 8\% reduction in parameters and at least 20\% reduction in FLOPs compared to DiT/SiT.  This demonstrates that our FlowDCN surpasses vision transformer-based generation models in multiple aspects.

\paragraph{Comparison with other generative models.} We report the final metrics of FlowDCN-XL/2 at \cref{tab:sota}. Our FlowDCN achieves much faster convergence speed with nearly $\frac{1}{5}$ total images compared its \textit{No-Long-residuals} counterparts. Additionally, using Euler ODE solver and classifier-free guidance with 1.375, our FlowDCN obtains SoTA 4.30 sFID and 2.13 FID results. Training for extra 400k steps, FlowDCN will be further improved to 2.00 FID. As sFID reflects the spatial structure quality~\cite{sfid}, better sFID shows our FlowDCN captures better structure distributions. We notice that the IS metric is lower than other models, however, there is an improvement trend along with training iterations.

\begin{figure}
    \vspace{-1em}
    \includegraphics[width=0.98\linewidth]{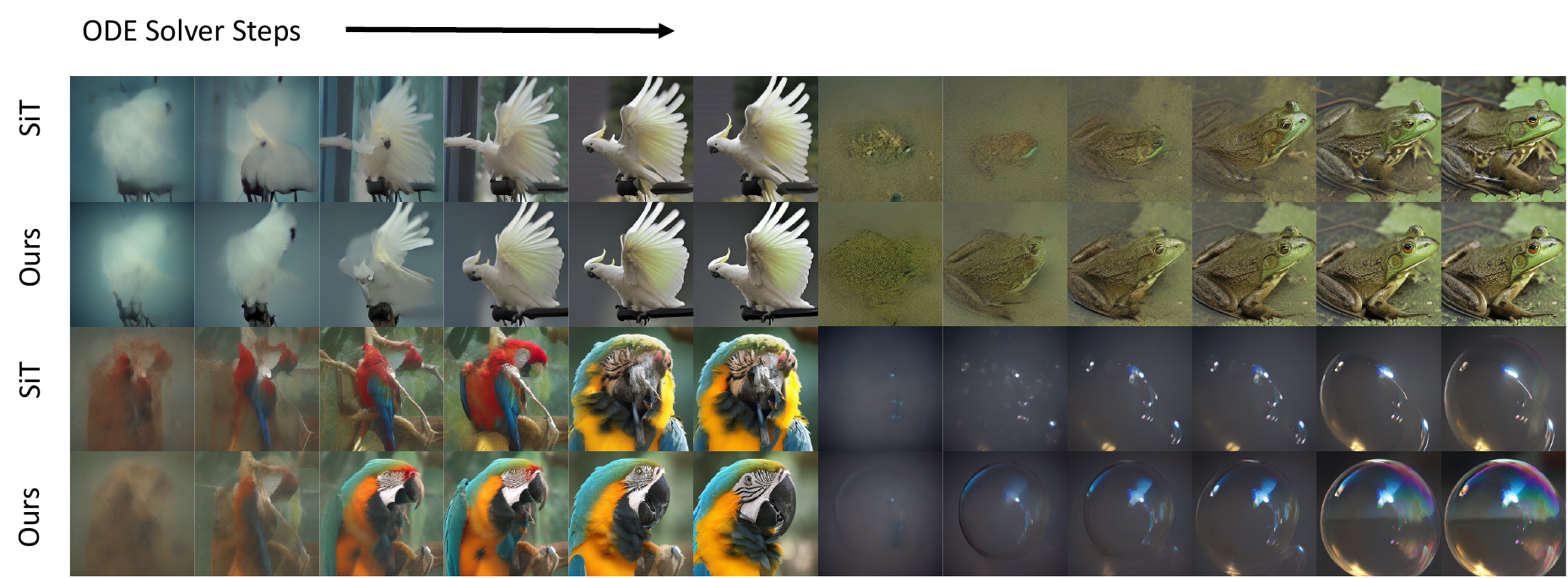}
    \caption{\textbf{Visualization Comparison with SiT}. \textit{Best viewed zoomed-in}. {\small We sample both our FlowDCN-XL/2 and SiT-XL/2 with Euler ODE solver under 2, 3, 4, 5, 8, 10 steps using the same latent noise. At the fewer steps sampling scenery, our FlowDCN generates slightly clearer and higher-quality images.}}
    \label{fig:vis}
    \vspace{-0.5em}
\end{figure}

\paragraph{Visual quality comparison with baseline SiT.} We sample both our FlowDCN-XL/2 and SiT-XL/2 with Euler ODE solver for 2, 3, 4, 5, 8, 10 steps, employing the same latent noise for both models. Notably, at the fewer steps sampling scenario, our FlowDCN generates slightly clearer and higher-quality images. We place the generated images at \cref{fig:vis} and Appendix. 

\input{article/tables/sota}

\subsection{512 $\times$ 512 ImageNet Dataset}
As training on high-resolution images consumes much more resources, we opt to fine-tune 100k steps from the same model trained on 256 × 256 resolution setting of 1.5M steps (corresponding to FlowDCN with 384M training images of \cref{tab:sota}). Although fine-tuned with limited 100k steps, our FlowDCN demonstrated powerful performance. 

\paragraph{Comparison with other generative models.} We report the final metrics of FlowDCN-XL/2 on 512 $\times$ 512 ImageNet Dataset at \cref{tab:sota512}. Our FlowDCN achieves much better FID and sFID performance compared to its counterparts. Using Euler SDE solver with 250 steps and classifier-free guidance with 1.375, our FlowDCN obtains 4.53 sFID and 2.44 FID results. Using Euler ODE solver with 50 steps and classifier-free guidance with 1.375, our FlowDCN obtains 5.29 sFID and 2.76 FID result. As shown in \cref{tab:sota512}, our FlowDCN achieves better sFID and captures better spatial structure distributions. 
\input{article/tables/imagenet512}

\subsection{Arbitrary Resolution Extension}
\begin{figure}
    \captionsetup[subfigure]{labelformat=empty}
   \subfloat[with $S_\text{max}$ Adjustment]{
        \centering
        \includegraphics[width=0.5\linewidth]{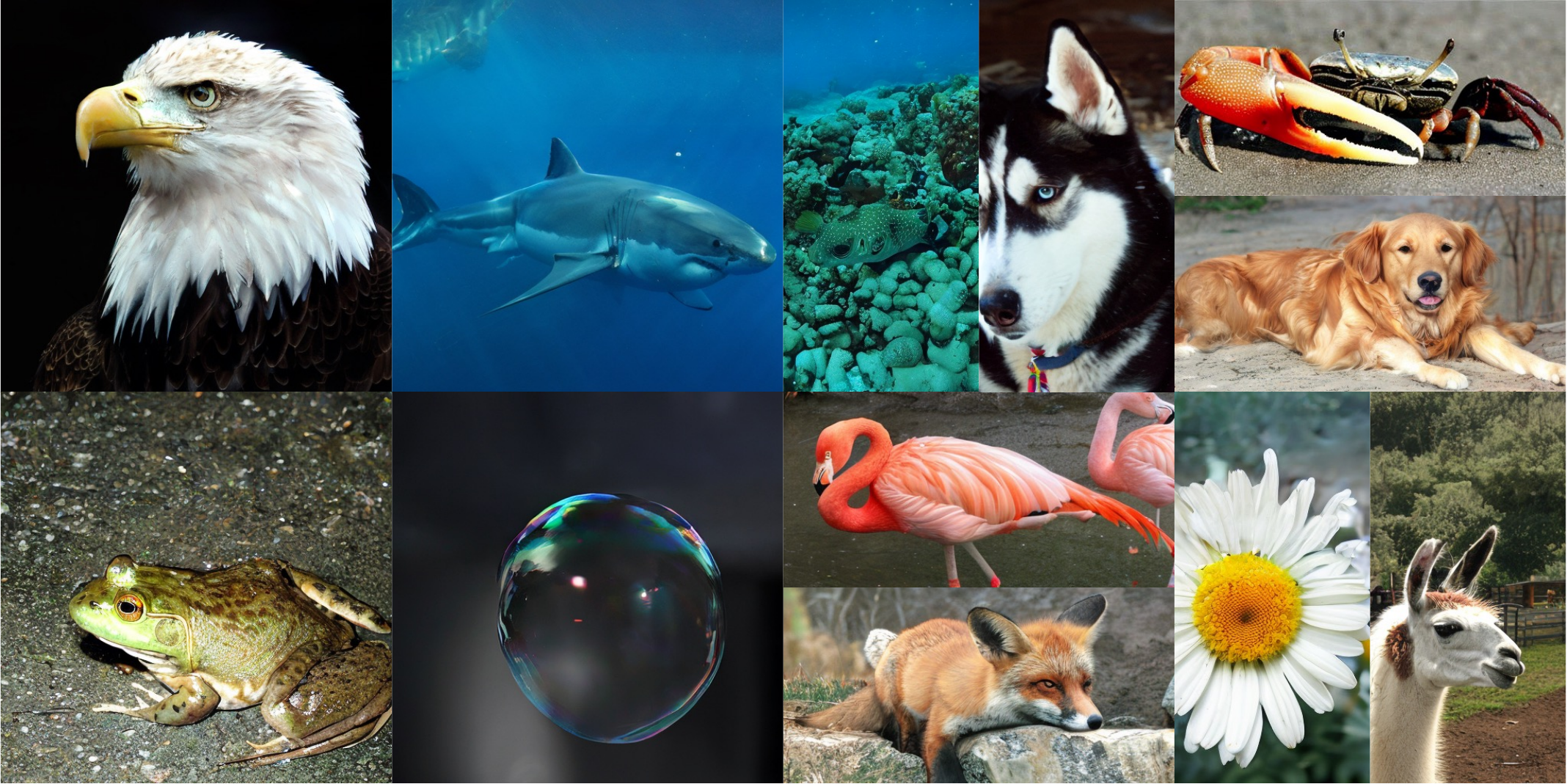}
    }
    \hspace{-2.1mm}
    \subfloat[without $S_\text{max}$ Adjustment]{
        \centering
        \includegraphics[width=0.5\linewidth]{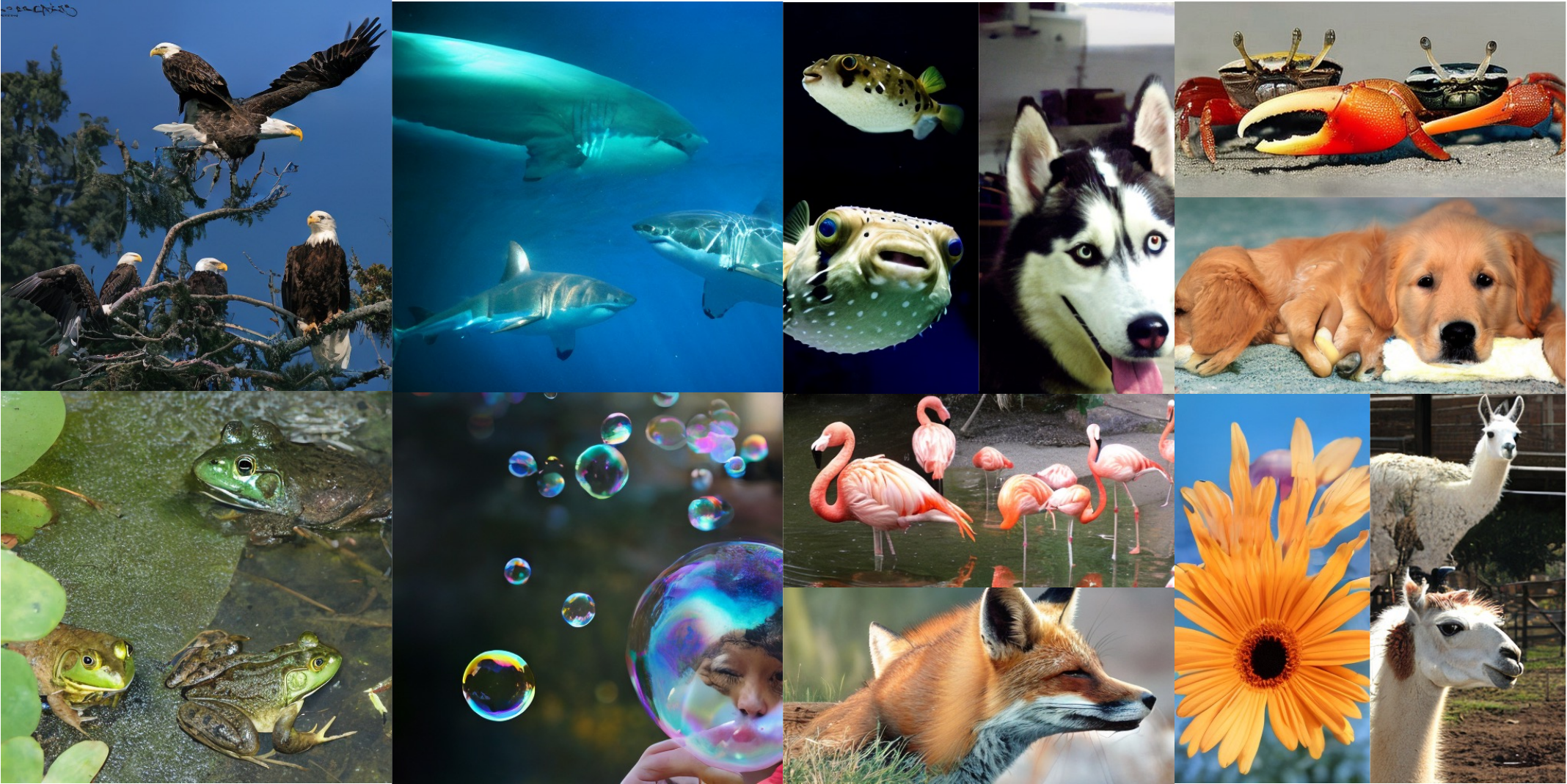}
    }%
    \caption{\textbf{Visualization Comparison about $S_\text{max}$ Adjustment.} {\small Here are the $512\times512$, $256\times512$ and $512\times256$, three type resolution images. We employ the same latent noise as start, sampling with Euler SDE solver for 250 steps.   With $S_\text{max}$ Adjustment, sampled images consistently looks better. }}
    \label{fig:vis_abs}
    \vspace{-1em}
\end{figure}
\input{article/tables/resolution}
For the resolution extrapolation evaluation, we follow the setting in FiT. We select 320x320 and 224x448 as the evaluation arbitrary resolution. It is worth noting that our FlowDCN can handle arbitrary resolution within a reasonable range, the reasonable range is determined by the training setting and training dataset. As our primary goal is to explore DCN-like architectures in universal image generation, we do not intend to enhance the resolution extrapolation nature by data processing. Therefore, we do not employ any multiple aspect ratio training techniques like FiT\cite{fit}. Instead, we directly use the FlowDCN model trained on the center-cropped 256x256 ImageNet dataset for arbitrary resolution extension experiments, showcasing the model's inherent capabilities. Moreover, we provide resolution extension experiments with various aspect ratio training techniques in the Appendix.

\textbf{Metric comparsion.} We report the evaluation results on \cref{tab:resolutions}. For Base-size models, our FlowDCN-B/2 achieves much better results on 320x320 resolution, with 34.4 FID and 35.7 FID using $S_\text{max}$ adjustment, outperforming FiT and DiT with a large margin.  On 224x448 resolution, our FlowDCN-B/2 achieves comparable results. Note our FlowDCN not employs any various aspect ratio training in \cref{tab:resolutions}, so we believe our FlowDCN-B/2 can achieve better results when incorporating such training augmentations. For large-size models, we report our FlowDCN-L/2 and FlowDCN-XL/2 with $S_\text{max}$ adjustment in \cref{tab:resolution_large}, our model shows comparable results to SoTA models. Meanwhile, we notice that FiT performs poorly on 256x256 resolution in \cref{tab:sota}, which we hypothesize is due to resolution-related data augmentation hurting the fitting power of original resolution distributions. Furthermore, as FiT employs the $256\times256$ reference statistics from ADM Eval Suite\cite{adm} to evaluate all resolution(even for $224\times448$), we suspect this evaluation paradigm is unreasonable. 

\textbf{Visual quality comparison of $S_\text{max}$.} In \cref{tab:resolution_base}, We notice FlowDCN-B/2 with $S_\text{max}$ adjustment does not exhibit better results than directly generating images, we hypothesize that FID and sFID are low level visual quality assessments, not reflecting semantic visual quality. So we also provide the visualization comparisons of our FlowDCN-XL/2 with and without $S_\text{max}$ Adjustment in \cref{fig:vis_abs} and Appendix. With $S_\text{max}$ Adjustment, generated images consistently look better. But not all the cases demand  $S_\text{max}$ Adjustment, some images like the bubble and the husky case in \cref{fig:vis_abs}, still look good even without $S_\text{max}$ Adjustment. More comparison examples can be found in the Appendix.

%% file: article/tables/op_speed.tex
\begin{table*}
    \centering
\resizebox{0.85\linewidth}{!}{
    \begin{tabular}{l|c|c|c|c}
    \shline
    \multicolumn{1}{c|}{\multirow{2}{*}{Operator}} & \multicolumn{4}{c}{Runtime (ms) of Input Shape $ H\times W \times G \times D $} \\ 
     & $16 \times 16 \times 16 \times 64 $ & $16 \times 16 \times 16 \times 128 $ & $32 \times 32 \times 16 \times 64$   & $32 \times 32 \times 16 \times 128 $\\
    \shline   
      Attention (Math SDP) & 0.92/2.1 & 1.16/2.71 & 10.7/28.8 & 12.4/35.8\\
      Attention (Flash SDP)~\cite{flashattention} & 0.62/N & 1.47/N & 4.98/N &14.4/N\\
      DeformConv(DCNv4~\cite{dcnv4}) & 0.77/1.00 & 1.0/2.1 & 2.8/4.4 & 3.9/8.3 \\
      DeformConv(Shm) & 0.56/0.81 & 1.1/1.5 & 2.7/3.9 & 5.0/7.3 \\
      DeformConv(Triton-lang)$\dagger$ & 0.83/0.89 & 0.95/1.1 & 3.4/3.8 & 4.0/4.8 \\
     \shline
    \end{tabular}
}

\caption{{\bf Op-level benchmark on standard input shape of Diffusion backbone task.}  {\small FP16/FP32 results are collected on Nvidia A10 GPU. We use 32 batch sizes for benchmarking. $\dagger$ indicates our Triton-lang~\cite{triton} implementation of DCNv4. N indicates implementation is not available.}
}
\label{tab:op_low_res}
\vspace{-1em} 
\end{table*}

%% file: article/tables/cifars.tex
\newcommand\tr{\rule{0pt}{10pt}}
\newcommand\br{\rule[-3.3pt]{0pt}{3.3pt}}

\begin{table*}[t]
\subfloat[
\textbf{Comparsions with SiT.}\label{tab:compare_cifar10}
{\small Our FlowDCN outperforms SiT by a significant margin.}
]{
\small
\centering
\begin{minipage}{0.32\linewidth}{\begin{center}
\resizebox{\linewidth}{!}{
\begin{tabular}{x{56}x{16}x{16}x{16}}
\shline
Models & FID$\downarrow$ & sFID$\downarrow$ & IS$\uparrow$ \\
\shline
SiT-S/2 & 7.42 & 4.47 & 8.7 \\
\hline
\textbf{FlowDCN-S/2} & \textbf{5.47} & \textbf{4.35} & \textbf{8.89} \\      
{ w/o MultScale} & {5.72} & {4.42} & {8.85}\\ 
{ w/o PriorInit} & 5.68 & 4.49 & 8.9 \\      
\shline
\end{tabular}}
\end{center}}
\end{minipage}
}
\hspace{0.5em}
\subfloat[
\textbf{\textit{KernelSize $K$} of FlowDCN.}\label{tab:kernel_size}
{\small large kernel size produces better results than small one.}
]{
\small
\centering

\begin{minipage}{0.3\linewidth}{
\begin{center}
\resizebox{0.97\linewidth}{!}{
\begin{tabular}{x{40}x{16}x{16}x{16}}
\shline
Kernel & FID$\downarrow$ & sFID$\downarrow$ & IS$\uparrow$ \\
\shline
4 & 5.88 &4.6 & 8.89 \\
\textbf{9} & \textbf{5.47} & \textbf{4.35} & \textbf{8.89}\\
16 & 5.39 & 4.54 &8.93\\
32 & 5.13 & 4.43 & 9.05\\
\shline
\end{tabular}}
\end{center}}
\end{minipage}
}
\hspace{.5em}
\subfloat[
\textbf{\textit{Deformable fields learning} setting.}\label{tab:priors}
{Default achieves best results.}
]{
\small
\centering
\begin{minipage}{0.3\linewidth}{\begin{center}
\resizebox{0.98\linewidth}{!}{
\begin{tabular}{x{20}x{20}x{12}x{12}x{12}}
\shline
$p_k$ & $s(\mathbf{x})$ & FID$\downarrow$ & sFID$\downarrow$ & IS$\uparrow$ \\
\shline
fixed & fixed & 5.6 & 4.58 & 8.90 \\
\textbf{fixed} & \textbf{learn} & \textbf{5.47} & \textbf{4.35} & \textbf{8.89} \\
learn & fixed & 6.01 & 4.43 & 8.85\\
learn & learn & 5.63 & 4.37 & 8.89\\ 
\shline
\end{tabular}}
\end{center}}
\end{minipage}
}
\caption{\textbf{Ablation Studies and Comprasion with other flow-based method on 32x32 CIFAR Dataset.} {\small In order to fully align with SiT~\cite{sit}, here we replace our SwiGLU and RMSNorm with FFN and LayerNorm armed in SiT, respectively. \textbf{Bold} font indicates the default setting.}} 
\vspace{-1em}
\end{table*}

%% file: article/tables/imagenet256.tex
\begin{table*}[ht]
\centering
\setlength{\tabcolsep}{2pt}
\small 
\begin{tabular}{*{1}{l}*{2}{l}*{10}{c}}
\toprule
Model & FLOPs (G) & Params (M) & Latency(ms) & FID$\downarrow$& sFID$\downarrow$  &IS$\uparrow$\\
\midrule
SiT-S/2 & 6.06 &33 & 0.026 & 57.64 &  9.05 &  24.78 \\
SiT-S/2 $^\dagger$ & 6.06 &33 & 0.026 & 57.9 & \textbf{8.72} & 24.64\\
\textbf{FlowDCN-S/2} & 4.36~({\color{ForestGreen} -28\%}) & 30.3~({\color{ForestGreen} -8.1\%})& 0.027 & \textbf{54.6} & 8.8 & \textbf{26.4} \\
\midrule
SiT-B/2 & 23.01 & 130 & 0.084 & 33.5 &  6.46 &  43.71 \\
SiT-B/2 $^\dagger$ & 23.01 & 130 & 0.084 & 37.3 & 6.55 & 40.6\\
\textbf{FlowDCN-B/2} & 17.87~({\color{ForestGreen} -22\%}) & 120~({\color{ForestGreen} -7.6\%}) & 0.076 & \textbf{28.5} & \textbf{6.09} & \textbf{51} \\ 
{\color{gray}w/o RMS \& SwiGLU} & 17.88~({\color{gray} -22\%}) & {\color{gray}120~(-7.6\%)} & {\color{gray}0.072} & {\color{gray}29.1} & {\color{gray}6.13} & {\color{gray}50.4}\\
\midrule
DiT-L/2 & 80.71& 458 & 0.291 &23.3&- &- \\
SiT-L/2 & 80.71 & 458 & 0.291 & 18.8 &  5.29 &  72.02 \\
\textbf{FlowDCN-L/2} & 63.51~({\color{ForestGreen} -21\%}) & 421~({\color{ForestGreen} -8.0\%}) & 0.254 & \textbf{13.8} & \textbf{4.69} & \textbf{85} \\
\midrule
DiT-XL/2 & 118.64 &675 & 0.387 & 19.5 &  - &  - \\
SiT-XL/2 & 118.64 &675 & 0.387 & 17.2 &  5.07 &  76.52 \\
\textbf{FlowDCN-XL/2} & 93.24~({\color{ForestGreen} -21\%}) & 618~({\color{ForestGreen} -8.4\%}) & 0.303 & \textbf{11.3} & \textbf{4.85} & \textbf{97} \\

\bottomrule
\end{tabular}
\caption{\textbf{Image generation metrics comparisons between SiT~\cite{sit}, DiT~\cite{dit} under 400k training steps budgets}. {\small All metrics are calculated from the sampled 50k images under 250 Euler SDE sampling steps without classifier-free guidance. $\dagger$: reproduced result. Latency(ms) is the 1-NFE latency and collected from Nvidia A10 GPU with 16 batchsize under float32.}}
\label{tab:sit_sde}
\vspace{-2em}
\end{table*}

%% file: article/tables/sota.tex
\begin{table*}[t]
\centering
\setlength{\tabcolsep}{2pt}
\resizebox{0.92\linewidth}{!}{
\begin{tabular}{l|ccc|ccccc}
\toprule
\multicolumn{5}{l}{\textbf{ImageNet 256×256 Benchmark}} \\
\midrule
\multirow{2}*{Generative Models}  & Long & Total & Total  & FID $\downarrow$ & sFID $\downarrow$& IS $\uparrow$& P $\uparrow$ & R $\uparrow$ \\
& Residuals & Images(M) & GFLOPs  &  &  &  &  &  \\
\midrule
ADM-U~\cite{adm} & \checkmark  & 507 &   $3.76\times10^{11}$  &     7.49  & 5.13 &   127.49   &  {0.72} &     0.63   \\
CDM~\cite{cdm} & \checkmark & - & - & 4.88  &   -  & {158.71}  &   -  &     -       \\
LDM-4~\cite{ldm} & \checkmark &  213 & $2.22\times10^{10}$ &  10.56   &  - &    103.49    & 0.71  &  0.62        \\
\hline
DiT-XL/2~\cite{dit} & \ding{55} & 1792 &  $2.13\times10^{11}$  & 9.62  &  6.85   &  121.50    & 0.67 &   \textbf{0.67}       \\
\arrayrulecolor{black}
DiffusionSSM-XL\cite{diffusionssm} & \ding{55} & 660 & $1.85\times10^{11}$ &  {9.07}   &  5.52 &   118.32   & 0.69    &  0.64       \\
SiT-XL/2\cite{sit} & \ding{55} & 1792 & $2.13\times10^{11}$ & 8.61 & 6.32 & \textbf{131.65} & 0.68 & \textbf{0.67} \\
\textbf{FlowDCN-XL/2} & \ding{55} & \textbf{384} & {$\mathbf{3.57\times10^{10}}$} & \textbf{8.36} & \textbf{5.39} &  {122.5 }& \textbf{0.69} & {0.65} \\
\bottomrule
\toprule
Classifier-free Guidance \\
\midrule
ADM-U\cite{adm}  & \checkmark & 507 & $3.76\times10^{12}$ &  3.60  &   -   & 247.67 & 0.87   & 0.48       \\
LDM-4~\cite{ldm} & \checkmark &  213 &  $2.22\times10^{10}$  &  3.95 &   -   &   178.22 & 0.81   &  0.55      \\
U-ViT-H/2~ \cite{uvit} & \checkmark  &  512  & $6.81\times10^{10} $ & 2.29  &  -   & 247.67 & {0.87}   & 0.48       \\
\hline
DiT-XL/2~\cite{dit} & \ding{55} & 1792 &   $2.13\times10^{11}$  &  2.27 &   4.60   & \textbf{278.24}  & 0.83  &  0.57    \\
DiffusionSSM-XL~\cite{diffusionssm} & \ding{55}&  660 &  $1.85\times10^{11}$  & 2.28 &  4.49     & 259.13  & \textbf{0.86}  &  0.56    \\
SiT-XL/2\cite{sit} & \ding{55} & 1792 & $2.13\times10^{11}$ &  {2.06} & 4.50 & 270.27 & 0.82 & \textbf{0.59} \\
FiT-XL/2\cite{fit} & \ding{55} & 450 & - &  4.27 & 9.99 & 249.72 & 0.84 & 0.51 \\
\textbf{FlowDCN-XL/2 {\small (cfg=1.375; ODE)}} & \ding{55} & \textbf{384} & {$\mathbf{3.57\times10^{10}}$} & {2.13} & \textbf{4.30} & {243.46} & {0.81} & {0.57}\\
\textbf{FlowDCN-XL/2 {\small (cfg=1.375; SDE)}} & \ding{55} & \textbf{384} & {$\mathbf{3.57\times10^{10}}$} & {2.08} & \textbf{4.38} & {257.53} & {0.82} & {0.57}\\
\textbf{FlowDCN-XL/2 {\small (cfg=1.375; ODE)}} & \ding{55} & \textbf{486} & {$\mathbf{4.52\times10^{10}}$} & \textbf{2.01} & \textbf{4.33} & {254.36} & {0.81} & {0.58}\\
\textbf{FlowDCN-XL/2 {\small (cfg=1.375; SDE)}} & \ding{55} & \textbf{486} & {$\mathbf{4.52\times10^{10}}$} & \textbf{2.00} & \textbf{4.37} & {263.16} & {0.82} & {0.58}\\
\bottomrule
\end{tabular}}
\caption{\textbf{Image generation quality evaluation of and existing approaches on ImageNet 256$\times$ 256}. {\small Total images by training steps $\times$ batch size as reported, and total GFLOPs by Total Images $\times$ GFLOPs/Image. P refers to Precision and R refers to Recall. 
}\label{tab:sota}}
\vspace{-1.5em}
\end{table*}

%% file: article/tables/imagenet512.tex
\begin{table}[t] 
    \begin{center}
    \begin{small}
    \scalebox{0.99}{
    \begin{tabular}{lccccc}
    \toprule
    \multicolumn{6}{l}{\bf{Class-Conditional ImageNet} 512$\times$512} \\
    \toprule
    Model & FID$\downarrow$   & sFID$\downarrow$  & IS$\uparrow$     & Precision$\uparrow$ & Recall$\uparrow$ \\
    \bottomrule
    BigGAN-deep~\cite{largegan} & 8.43 & 8.13 & 177.90 & 0.88 & 0.29 \\
    StyleGAN-XL~\cite{styleganxl} & 2.41 & 4.06 & 267.75 & 0.77 & 0.52 \\
    \toprule
    ADM~\cite{adm} & 23.24 & 10.19 & 58.06 & 0.73 & 0.60  \\
    ADM-U~\cite{adm} & 9.96 & 5.62 & 121.78 & 0.75 & {0.64} \\
    ADM-G~\cite{adm} & 7.72 & 6.57 & 172.71 & {0.87} & 0.42 \\
    ADM-G, ADM-U & 3.85 & 5.86 & 221.72 & 0.84 & 0.53\\
    {DiT-XL/2}~\cite{dit} & 12.03 & 7.12 & 105.25 & 0.75 & {0.64} \\
    {DiT-XL/2-G}~\cite{dit} (cfg=1.50) & {3.04} & {5.02} & {240.82} & 0.84 & 0.54 \\
    {SiT-XL/2-G} ~\cite{sit} (cfg=1.50) & 2.62 &  4.18 & 252.21 & 0.84 & 0.57 \\
    \textbf{FlowDCN-XL/2(cfg=1.375, ODE-50)} & 2.76 & 5.29 & 240.6 & 0.83 & 0.51 \\
    \textbf{FlowDCN-XL/2(cfg=1.375, SDE-250)} & \textbf{2.44} & \textbf{4.53} & \textbf{252.8} & 0.84 & 0.54 \\
    \bottomrule
    \end{tabular}
    } %
    \end{small}
    \end{center}
    \caption{\textbf{Benchmarking class-conditional image generation on ImageNet 512$\times$512.} {\small Our FlowDCN-XL/2 is fine-tuned for 100k steps from the same model trained on $256\times256$ resolution setting of 1.5M steps}}
    \label{tab:sota512}
    \vspace{-2em}
\end{table}

%% file: article/tables/resolution.tex
\begin{table*}
\subfloat[
\textbf{Metrics Results on Base-Size Models}\label{tab:resolution_base}
]{
\centering
\begin{minipage}{0.48\linewidth}{\begin{center}
\resizebox{0.95\linewidth}{!}{
\begin{tabular}{lx{16}x{16}x{16}|x{16}x{16}x{16}}
\toprule
\multirow{2}*{Method}&\multicolumn{3}{c|}{320$\times$320 (1:1)}&\multicolumn{3}{c}{224$\times$448 (1:2)} \\

& {FID$\downarrow$} & {sFID$\downarrow$} & {IS$\uparrow$}
& {FID$\downarrow$} & {sFID$\downarrow$} & {IS$\uparrow$} \\ 

\midrule
DiT-B & 95.5 & 108.7 & 18.4  &     
109.1 & 110.7 & 14.0  \\   

DiT-B$_\text{EI}$ & 81.5 & 62.3 & 21.0 &    
133.2 & 72.5 & 11.1  \\

DiT-B$_\text{PI}$ & 72.5 & 54.0 & 24.2 &    
133.4 & 70.3 & 11.7 \\  

\midrule
FiT-B & 61.4 & {30.7} & 31.0 & 44.7 & \textbf{24.1} & 37.1 \\
FiT-B$_\text{vYaRN}$ & {44.8} & 38.0 & {44.7} & \textbf{41.9} & 42.8 & \textbf{45.9}  \\
FiT-B$_\text{vNTK}$ & 57.3 & 31.3 & 34.0 & 43.8 & 26.3 & 39.2 \\
\midrule
\textbf{FlowDCN-B/2} & \textbf{34.4} & \textbf{27.2} & \textbf{52.2} & 71.7 & 62.0 & 23.7\\
\textbf{\small + $S_\text{max}$ Adjust} & \textbf{35.7} & \textbf{29.3} & \textbf{51.2} & 81.1 & 40.2 & 21.1\\
\bottomrule
\end{tabular}}
\end{center}}
\end{minipage}
}
\hspace{.5em}
\subfloat[
\textbf{Metrics Results on Large-Size Models}\label{tab:resolution_large}
]{
\centering
\begin{minipage}{0.48\linewidth}{\begin{center}
\resizebox{0.98\linewidth}{!}{
\begin{tabular}{lx{16}x{16}x{16}|x{16}x{16}x{16}}
\toprule
\multirow{2}*{Method}&\multicolumn{3}{c|}{320$\times$320 (1:1)}&\multicolumn{3}{c}{224$\times$448 (1:2)} \\

& {FID$\downarrow$} & {sFID$\downarrow$} & {IS$\uparrow$}
& {FID$\downarrow$} & {sFID$\downarrow$} & {IS$\uparrow$} \\ 
\midrule
{ADM-G,U}~\cite{adm} & 9.39 & \textbf{9.01} & 162 & 11.34 & \textbf{14.5} & 146 \\
LDM-4~\cite{ldm} & 6.24 & 13.21 & 220 & 8.55 & 17.62 & 186 \\
\midrule
UViT-H/2~\cite{uvit} & 7.65 & 16.30 & 208 & 67.1 & 42.92 & 45.5 \\
MDT-G~\cite{mdt} & 383 & 136 & 4.24 & 365 & 142.8 & 4.91 \\
DiT-XL/2~\cite{dit} & 9.98 & 23.57 & 225 & 94.94 & 56.06 & 35.7 \\
FiT-XL/2~\cite{fit}& \textbf{5.42} & 15.41 & 252& \textbf{7.9} & 19.63 & \textbf{215} \\
\midrule
\textbf{FlowDCN-L/2} & 5.99 & 9.71 & 238 & 12.8 & 17.9 & 168 \\
\textbf{FlowDCN-XL/2} & 5.86 & 13.5 & \textbf{275} & 12.9 & 20.6 & 184 \\
\bottomrule
\end{tabular}}
\end{center}}
\end{minipage}
}
\caption{\textbf{Benchmarking resolution extrapolations on ImageNet dataset.} { \small On the Base-size Models benchmark, our FlowDCN achieves much better results on 320x320 resolution and comparable results on 224x448  resolution. On the Large-Size Models benchmark, our flowDCN shows comparable extrapolation performance to SoTA models}.\label{tab:resolutions}}
\vspace{-2em}
\end{table*}

%% file: article/conclusion.tex
\section{Conclusion}
In this paper, we have presented FlowDCN, a novel deformable convolutional network for arbitrary-resolution image generation. Our FlowDCN model leverages the strengths of both group-wise multiscale deformable convolutions and linear flow to generate high-quality images of various resolutions with high flexibility. Through extensive experiments, we demonstrate that FlowDCN outperforms the state-of-the-art transformer-based counterparts in terms of performance, convergence speed, and computational efficiency. Additionally, our model exhibits strong resolution extrapolation capabilities, achieving comparable results to previous models on arbitrary resolution without any additional training techniques. We believe that FlowDCN has a great potential to become a powerful tool for a wide range of image generation tasks and applications.
\vspace{-1em}
\section*{Limitations and Broader Impacts}
\vspace{-1em}
Our current implementation of deformable convolution backward is inefficient to be on par with Attention. Our primary focus remains on optimizing the training speed. Once we have made significant strides in training optimization, we plan to scale up our FlowDCN to accommodate larger model parameters and higher training resolution, paving the way for more advanced explorations.
\vspace{-1em}
\section*{Acknowledgement}
\vspace{-1em}
This work is supported by the National Key R$\&$D Program of China (No. 2022ZD0160900), the National Natural Science Foundation of China (No. 62076119), the Fundamental Research Funds for the Central Universities (No. 020214380119), the Nanjing University-China Mobile Communications Group Co., Ltd. Joint Institute, and the Collaborative Innovation Center of Novel Software Technology and Industrialization.

%% file: article/supp1.tex
\section*{A. ~Model Details}
\begin{table}[H]
\centering
\small
\begin{tabular}{l c c c c c}
\toprule
Model            & Layers $N$ & Hidden size $d$ &  Groups   \\
\midrule 
FlowDCN-S   &   12   &     384   &   6    \\
FlowDCN-B   &   12   &      768    &   12    \\
FlowDCN-L  &    24   &      1024    &   16    \\
FlowDCN-XL &    28  &       1152     &   16   \\
\bottomrule
\end{tabular}
\caption{\textbf{Details of FlowDCN models.} {\small We follow DiT for the Small (S), Base (B), Large (L) and XLarge (XL) model configurations.}}

\label{tbl:models}
\end{table}

\section*{B. Comparisons between FlowCNN and FlowDCN on ImageNet $256\times256$}
\textbf{The relationship between DCN and common CNN.} 
As \cref{eq:dcn} states, DCN introduces a deformable field $\Delta p(x)$ and dynamic weight $w(x)$. When all features shares the same static weight instead of dynamic, and deformable field $\Delta p(x)$ degrades to zeros, DCN degenerates to common CNN.Therefore, in most scenarios, DCN-like architectures are more powerful than common CNNs. Furthermore, the \textit{fix $p_k$} in \cref{tab:priors}indicates that we freeze the $p_k$ (not the deformable field $\Delta p(x)$) and initialize it with a predefined grid.

\textbf{Why not try a common CNN architecture.} In many computer vision tasks, traditional CNNs have been outperformed by transformers, so we opted to explore the modern, advanced CNN variant, Deformable Convolutional Networks (DCN). Additionally, we conducted a small experiment where we replaced the DCN block in FlowDCN with standard 3x3 and 5x5 group-wise convolution blocks.

\begin{table*}[ht]
\centering
\setlength{\tabcolsep}{2pt}
\small 
\begin{tabular}{*{1}{l}*{7}{c}}
\toprule
Model	& layers&	groups&	channels&	Params (M)&	FID&	sFID	IS\\
\toprule
SiT-S/2&	12 &	6&	384&	33.0&	57.64&	9.05&	24.78\\
FlowCNN-3x3	&12&	8&	512&	49.1&	59.0&	10.7&	27.4\\
FlowCNN-5x5	&12&	6&	384&	33.1&	63.0&	10.9&	23.6\\
FlowDCN-S/2	&12&	6&	384&	30.3&	54.6&	8.8&	26.4\\
\bottomrule
\end{tabular}
\caption{\textbf{Image generation metrics comparisons between SiT, FlowDCN and FlowCNN under 400k training steps budgets}.}
\label{tab:sit_sde}
\end{table*}
\input{article/tables/resolution_var}

\section*{C. ~ Resolution Extension with Various Aspect Ratios Training}
While FlowDCN, trained on fixed-resolution images, is capable of generating images of arbitrary resolution within a reasonable aspect ratio range, its performance can be improved by adopting variable aspect ratio (VAR) training instead of a fixed 256x256 resolution. To ensure a fair comparison with FiT, which inherently uses VAR, we train a FlowDCN-B/2 model from scratch using VAR techniques. We evaluate our model using the same pipeline and reference batch as FiT, without CFG.





    


%% file: article/tables/resolution_var.tex
\begin{table}[ht]
\centering
\begin{adjustbox}{max width=1.\textwidth}
\begin{tabular}{l|ccc|ccc|ccc|ccc}
\toprule[1.2pt]
\multirow{2}*{Method}&\multicolumn{3}{c|}{256$\times$256 (1:1)}&\multicolumn{3}{c|}{320$\times$320 (1:1)}&\multicolumn{3}{c|}{224$\times$448 (1:2)}&\multicolumn{3}{c}{160$\times$480 (1:3)} \\

& \textbf{FID$\downarrow$} & \textbf{sFID$\downarrow$} & \textbf{IS$\uparrow$}
& \textbf{FID$\downarrow$} & \textbf{sFID$\downarrow$} & \textbf{IS$\uparrow$} 
& \textbf{FID$\downarrow$} & \textbf{sFID$\downarrow$} & \textbf{IS$\uparrow$}
& \textbf{FID$\downarrow$} & \textbf{sFID$\downarrow$} & \textbf{IS$\uparrow$}   \\

\midrule

DiT-B &  44.83 & {8.49} & 32.05 & 95.47 & 108.68 & 18.38 &      109.1 & 110.71 & 14.00  &  143.8 & 122.81 & 8.93 \\

DiT-B + EI &  44.83 & {8.49} & 32.05 & 81.48 & 62.25 & 20.97 &     133.2 & 72.53 & 11.11 & 160.4 & 93.91 & 7.30 \\

DiT-B + PI &  44.83 & {8.49} & 32.05 & 72.47 & 54.02 & 24.15 & 133.4 & 70.29 & 11.73 & 156.5 & 93.80 & 7.80 \\
\midrule

FiT-B & {36.36} & 11.08 & {40.69} & 61.35 & {30.71} & 31.01 & 44.67 & {24.09} & 37.1 & 56.81 & {22.07} & 25.25 \\

FiT-B + {VisionYaRN}  & {36.36} & 11.08 & {40.69} & {44.76} & 38.04 & {44.70} & {41.92} & 42.79 & {45.87} & 62.84 & 44.82 & {27.84} \\

FiT-B + {VisionNTK}  & {36.36} & 11.08 & {40.69} & 57.31 & 31.31 & 33.97 & 43.84 & 26.25 & 39.22 & {56.76} & 24.18 & 26.40 \\

\midrule

FlowDCN-B & 28.5 & 6.09 & 51 & 34.4 & 27.2 & 52.2 & 71.7 & 62.0 & 23.7 & 211 & 111 & 5.83 \\
FlowDCN-B (+VAR)  & 23.6 & 7.72 & 62.8 & 29.1 & 15.8 & 69.5 & 31.4 & 17.0 & 62.4 & 44.7 & 17.8 & 35.8 \\
+ $S_\text{max}$ Adjust  & 23.6 & 7.72 & 62.8 & 30.7 & 19.4 & 68.5 & 37.8 & 22.8 & 54.4 & 53.3 & 22.6 & 31.5 \\
\bottomrule[1.2pt]
\end{tabular}
\end{adjustbox}
\vspace{2em}
\caption{ \textbf{Benchmarking resolution extrapolations on ImageNet with various aspect ratio training}. {\small VAR indicates various aspect ratios training. We follow the same evaluation pipeline of FiT without using CFG}.}
\label{tab:ablation_in1k_ood}
\end{table}